%% file: main.tex
\documentclass[10pt,twocolumn,letterpaper]{article}
\usepackage[pagenumbers]{cvpr}
\makeatletter
\let\GaussTransKernelAddToCurCol\@addtocurcol
\makeatother
\input{preamble}
\usepackage{needspace}
\input{authors}
\newif\ifGaussTransBodyAlignment
\makeatletter
\patchcmd{\cvprsection}{-10pt}{-8pt}{}{\PackageError{GaussTrans}{Section before-gap patch failed}{}}
\patchcmd{\cvprsection}{7pt}{5pt}{}{\PackageError{GaussTrans}{Section after-gap patch failed}{}}
\apptocmd{\normalsize}{\ifGaussTransBodyAlignment\baselineskip=12pt plus 0.1pt\relax\fi}{}{\PackageError{GaussTrans}{Body baseline alignment patch failed}{}}
\apptocmd{\abstract}{\baselineskip=12pt plus 0.1pt\relax}{}{\PackageError{GaussTrans}{Abstract baseline alignment patch failed}{}}
\makeatother
\AtBeginDocument{%
  \GaussTransBodyAlignmenttrue\normalsize
  \setlength{\abovedisplayskip}{8pt}%
  \setlength{\belowdisplayskip}{3pt}%
  \setlength{\abovedisplayshortskip}{2pt}%
  \setlength{\belowdisplayshortskip}{3pt}%
}
\definecolor{cvprblue}{rgb}{0.21,0.49,0.74}
\usepackage[breaklinks,colorlinks,allcolors=cvprblue]{hyperref}
\hypersetup{pdftitle={GaussTrans: A Gaussian Interface Framework for Multi-Layer Refractive Transport},pdfauthor={\GaussTransPDFAuthors}}

\title{GaussTrans: A Gaussian Interface Framework\\for Multi-Layer Refractive Transport}
\author{\GaussTransAuthors}
\makeatletter
\g@addto@macro\@maketitle{\vspace*{-18pt}\input{figures/teaser}\vspace*{6pt}}
\makeatother

\begin{document}

\flushbottom
\captionsetup{skip=6pt}
\maketitle
\captionsetup{skip=4pt}
\input{sections/00_abstract}
\input{sections/01_introduction}
\input{sections/02_related_work}
\input{sections/03_method}

\input{sections/04_experiments}
\input{sections/05_conclusion}
\par
\GaussTransBodyAlignmentfalse\normalsize

{\small
\bibliographystyle{ieeenat_fullname}
\bibliography{references}
}
\input{supplement_body}
\end{document}

%% file: preamble.tex
\usepackage{caption}
\usepackage{placeins}
\usepackage{tabularx}
\usepackage{microtype}
\AtBeginDocument{\frenchspacing}
\usepackage{flafter}
\makeatletter
\patchcmd{\cvprsection}{-10pt plus -2pt minus -2pt}{-10pt}{}{%
  \PackageError{GaussTrans}{Unable to fix section spacing}{}%
}
\patchcmd{\cvprsubsection}{-8pt plus -2pt minus -2pt}{-8pt}{}{%
  \PackageError{GaussTrans}{Unable to fix subsection spacing}{}%
}
\patchcmd{\cvprsubsubsection}{-6pt plus -2pt minus -2pt}{-6pt}{}{%
  \PackageError{GaussTrans}{Unable to fix subsubsection spacing}{}%
}
\renewcommand\paragraph{\@startsection{paragraph}{4}{\z@}%
  {8pt}{-1em}{\normalfont\normalsize\bfseries}}
\newcommand{\GaussTransCompactSection}[1]{\section{#1}}
\patchcmd{\endabstract}{\vspace*{12pt}}{\par}{}{%
  \PackageError{GaussTrans}{Unable to normalize abstract transition}{}%
}
\patchcmd{\abstract}{\vspace*{12pt}}{\vspace*{7pt}}{}{\PackageError{GaussTrans}{Abstract gap patch failed}{}}
\makeatother

\AtBeginDocument{%
  \setlength{\abovedisplayskip}{8pt}%
  \setlength{\belowdisplayskip}{3pt}%
  \setlength{\abovedisplayshortskip}{2pt}%
  \setlength{\belowdisplayshortskip}{3pt}%
  \setlength{\jot}{2pt}%
  \thickmuskip=5mu\relax
  \medmuskip=4mu\relax
  \predisplaypenalty=10000
}

%% file: authors.tex
\newcommand{\GaussTransAuthors}{%
Zixuan Zhao\qquad Tao Wang\textsuperscript{*}\\
Sichuan University\\
{\small\textsuperscript{*}Corresponding author.}}
\newcommand{\GaussTransPDFAuthors}{Zixuan Zhao; Tao Wang}

%% file: figures/teaser.tex
\begin{minipage}{\textwidth}
\centering
\includegraphics[width=\linewidth]{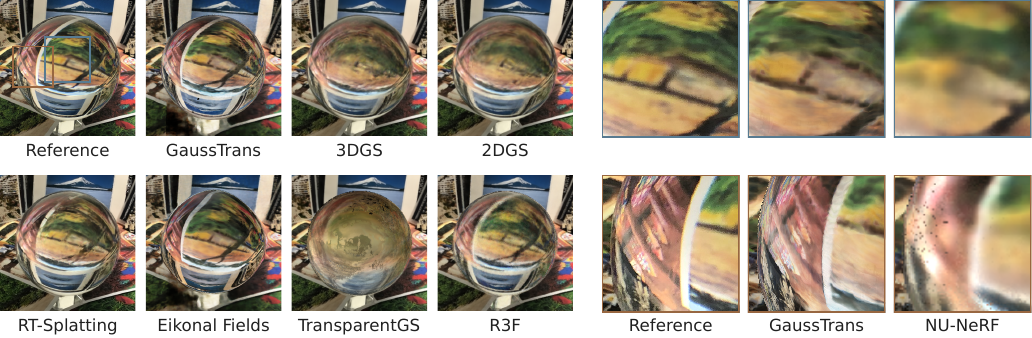}
\captionsetup{hypcap=false}
\captionof{figure}{Comparison on the real Ball scene~\cite{bemana2022eikonal}. Left: reference photograph and method comparisons. Right: matched close-ups of the boxed regions. Experimental details are provided in the supplementary material.}
\label{fig:teaser}
\captionsetup{hypcap=true}
\end{minipage}

%% file: sections/00_abstract.tex
\begin{abstract}
We investigate surface modeling of multi-transparent interfaces to achieve accurate light transmission. The interplay of errors in geometric structure, light paths, and observation results during the reconstruction of transparent media with a high number of layers makes this a challenging problem. Some existing techniques address the reconstruction of transparent scenes by reducing the number of layers or simplifying light paths; however, this limits their direct applicability to high-order transparent scenes. We propose GaussTrans, a compact framework for light transport in high-order transparent scenes, consisting of three components: Gaussian-Field Hybrid Surfaces, Transport-Coupled Refinement, and Primitive-Guided Implicit Intersection. Gaussian-Field Hybrid Surfaces use a hybrid approach combining Gauss surfaces and fields to reconstruct different interfaces. Transport-Coupled Refinement reduces coupling errors by evaluating the entire light path. Primitive-Guided Implicit Intersection combines primitive support regions to reduce errors in the intersection of interfaces and rays. In experiments on datasets such as RefRef, GaussTrans achieves significantly higher accuracy with relatively low spatial storage requirements.
\end{abstract}

%% file: sections/01_introduction.tex
\par\penalty-10000\relax
\vspace*{-\baselineskip}
\section{Introduction}
\label{sec:introduction}
Reconstructing transparent objects has long been a challenge in 3D scene reconstruction, and high-order transparent scene reconstruction has found widespread applications in multi-element optical systems~\cite{pharr2023pbrt}, transparent objects with high overlap angles, and experimental glassware~\cite{li2025tsgs}. However, as the number of layers increases, the nature of this problem fundamentally changes: errors in the position or normal of upstream interfaces can alter subsequent intersection points and incident directions, causing local approximation errors to couple along the optical path. This poses challenges for geometric accuracy and surface correction strategies. Some reconstruction and rendering methods restrict light paths to two interface interactions~\cite{li2020lookingglass,chen2023nerrf} or model only transparent horizontal planes~\cite{yoon2026wsgs}, thereby limiting their direct applicability to higher-order light paths. Other differentiable rendering frameworks support recursive or multiple-interaction light propagation~\cite{vicini2021pathreplay,jakob2022drjit,zhang2026radiancecache}. However, tracking more interactions does not in itself automatically eliminate geometric error coupling between different interfaces. In addition, some methods address transparent-object imaging by separately modeling transmission, refraction, and reflection components~\cite{na2026glint,shi2026rtsplatting}. However, extending such formulations to multilayer media can substantially increase modeling and computational complexity.

\input{figures/pipeline_overview}
In this paper, we propose GaussTrans: a compact interface framework for high-order transparent scenes that combines Gaussian and field-value methods. We construct the interface field using Gaussian-weighted local polynomial surface patches, where the local polynomials describe curvature and the Gaussian weights organize their spatial influence; intersection points correspond to field values of zero, forming the transparent interface structure. The Transport-Coupled Refinement module further supports constrained joint geometric refinement by sharing deformation controls and coordinating refitting across multiple view-correlated backgrounds. Primitive-Guided Implicit Intersection directs the intersection search range without altering the field.

In summary, our contributions are as follows:
\begin{enumerate}[partopsep=0pt]
\item \textbf{Gaussian-Field Hybrid Surfaces:} Merging Gaussian-weighted \(N\)th-order local polynomial surface patches (third-order by default in this paper) into compact optical surfaces for multilayer refractive light transmission.
\item \textbf{Transport-Coupled Refinement:} Multi-view background correspondence observes common constraints on the local deformation parameters shared by each glass shell, enabling the coordinated update of Gaussian centers, local coordinate systems, and surface coefficients.
\item \textbf{Primitive-Guided Implicit Intersection:} Utilizing Gaussian support intervals and local polynomial roots, this method guides the search for candidate positions along the ray and validates intersection points.
\end{enumerate}

%% file: figures/pipeline_overview.tex
\begin{figure*}[t]
  \centering
  \includegraphics[width=\textwidth]{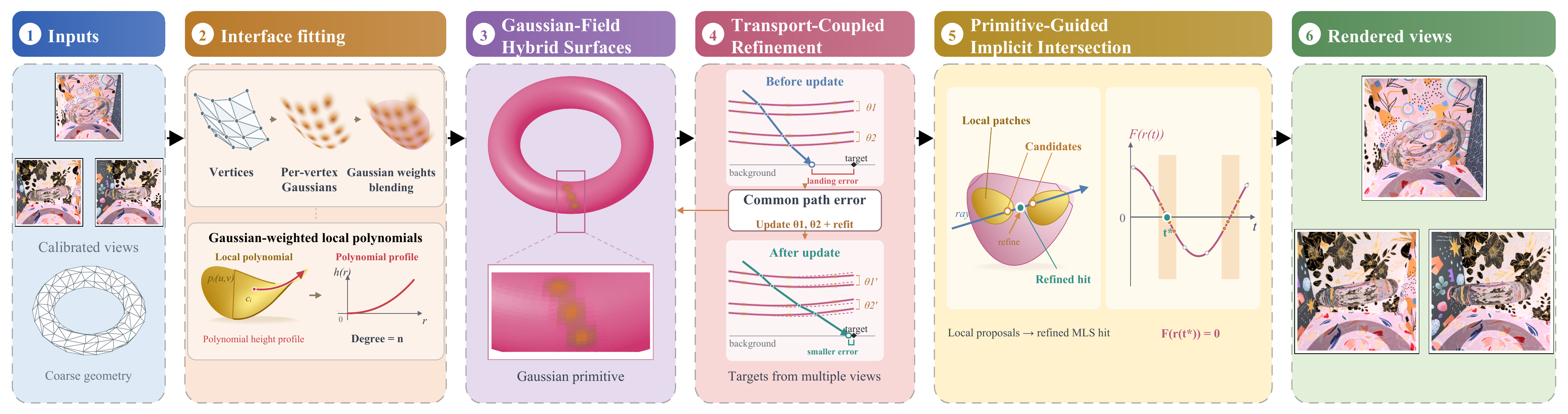}
  \caption{GaussTrans overview. (1--3) Inputs and local fitting construct Gaussian-Field Hybrid Surfaces. (4) Transport-Coupled Refinement updates interface geometry using multi-view path errors. (5) Primitive-Guided Implicit Intersection proposes and validates intersections. (6) Light transport through the interfaces produces rendered views.}
  \label{fig:pipeline}
\end{figure*}

%% file: sections/02_related_work.tex
\GaussTransCompactSection{Related Work}
\label{sec:related_work}
\subsection{Transparent Geometry Reconstruction}
Transparent geometric reconstruction has undergone significant development, primarily through the reconstruction of refraction constraints by reconstructing transparent interfaces. DRT~\cite{lyu2020drt} obtains ray--background mappings from the encoded background and optimizes the mesh using differentiable refraction tracing. NeTO~\cite{li2023neto} uses refractive constraints that account for self-occlusion to fit an implicit signed distance field. RCTrans~\cite{gao2025rctrans} combines confidence filtering to estimate ray--background correspondences in natural scenes. Among photometric methods, Gao et al.~\cite{gao2023implicitrefraction} estimate object contours from a reconstructed neural field, then optimize the neural SDF using differentiable refractive rendering supervised by image color. TNSR~\cite{deng2024tnsr}, proposed by Deng et al., starts from an initial implicit surface and, by taking derivatives of reflection and refraction traces, propagates multi-view RGB errors to surface parameters for refinement; its explicit tracing uses up to two refractions and one reflection. ReNeuS~\cite{tong2023reneus} reconstructs internal objects when the geometry of the transparent shell is known; NU-NeRF~\cite{sun2024nunerf} employs an outside-in reconstruction process. DiffTrans~\cite{li2026difftrans} jointly estimates geometry, refractive index, and absorption through recursive differentiable ray tracing. RefracGS~\cite{shao2026refracgs} jointly reconstructs a refractive water surface represented by a neural height field and the scene beneath it. In multi-interface scenes, increasing the number of interfaces can raise the costs of these methods or render them inapplicable. Method design should therefore jointly consider representation cost, applicability, and light-path accuracy.

\subsection{Gaussian Surfaces and Ray Intersection}
{\frenchspacing
Weighted local fits and implicit surface blending have established precedents in point-set and partition-of-unity surfaces~\cite{alexa2001pointset,ohtake2003mpu,guennebaud2007apss}. There have been relevant studies on Gaussian surfaces: 3DGS~\cite{kerbl2023gaussians} uses volumetric primitives but conflates appearance with geometry, while 2DGS~\cite{huang2024surfsplatting} and QGS~\cite{zhang2025qgs} use planes and quadratic surfaces, respectively. GOF~\cite{yu2024gof} extracts opacity surfaces; GSrec~\cite{wu2024gsrec} recovers surfaces from 3DGS using local structural hints. 3DGRT~\cite{moenneloccoz2024gaussianray} accelerates particle tracing. TSGS~\cite{li2025tsgs} is designed for first-surface extraction, but complex multi-layer refraction remains a limitation. TransparentGS~\cite{huang2025transparentgs} combines Gaussian with a mesh proxy, assuming two refractions and at most one total internal reflection. GLINT~\cite{na2026glint} decomposes the interface, transmission, and reflection components under the condition that refractive deflection is negligible. RT-Splatting~\cite{shi2026rtsplatting} combines surface and volumetric rendering to model reflection and transmission. Local polynomial intersections and support bounds also underlie earlier point-set ray tracers~\cite{adamson2003ray,adamson2003approx}. As a surface construction method, GaussTrans should also include ray-intersection capabilities: SuGaR~\cite{guedon2024sugar} determines the sampling interval based on the scale of the corresponding Gaussian along the line of sight; IRGS~\cite{gu2025irgs} traces secondary rays along two-dimensional Gaussians to query visibility and indirect incident radiance. GRay~\cite{poirierginter2026gray} uses smaller Gaussians to reduce overlap between primitives and organizes spatial queries using oriented bounding boxes. For continuous multi-interface refraction, local surface representations or efficient primitive queries alone are insufficient. Some methods still rely on limited interactions or negligible refraction, leaving reliable intersection finding throughout multi-layer refractive propagation an outstanding challenge.\par}

\subsection{Geometric Update Strategies}
Previous work has proposed various geometric update strategies tailored to different scene structures and interface geometries. SC-GS~\cite{huang2024scgs} predicts local rotation and translation for each control point, then weights the transformations of neighboring control points to drive Gaussian motion. Mango-GS~\cite{huang2026mangogs} associates Gaussians with sparse control nodes based on position and features, and transfers the temporal deformations of the nodes to the Gaussians. MoSca~\cite{lei2025mosca}, on the other hand, organizes sparse motion nodes into a graph and coordinates the motion of adjacent nodes through local rigidity constraints. Xu et al.~\cite{xu2025hybrid} model the vertex displacement field in patches, allowing vertices within the same region to be driven by a shared local deformation function. Ikeda et al.~\cite{ikeda2025thickness} use a neural signed distance field and a wall-thickness parameter to jointly describe the outer and inner interfaces. These methods coordinate local updates through shared deformations or geometric constraints. However, geometric coordination alone does not ensure consistent multi-interface refractive transport; the effects of updates to different interfaces on subsequent propagation and multi-view observations must also be considered jointly.

%% file: sections/03_method.tex
\section{Method}
\label{sec:method}
\input{figures/mls_mechanism}
\subsection{Overall Framework}
\label{sec:overview}
Our work primarily focuses on multi-layer transparent scenes. To balance generalization capability with the error coupling caused by multi-layer transparent interfaces, we adopt the following prior constraints:
Given rough interface geometry, known medium properties, a calibrated camera, and a background or environment model, GaussTrans first fits Gaussian-weighted local \(N\)th-order polynomial surface patches (cubic in this paper) and constructs a field for each interface. Subsequently, within a finite computational budget, it considers the propagation of light rays across the interfaces and predicts background landing positions or rendered images. The Transport-Coupled Refinement module provides a refinement strategy for multi-layer interfaces; Primitive-Guided Implicit Intersection, serving as the intersection module, improves the robustness of light-path--interface intersections. Figure~\ref{fig:pipeline} summarizes the overall workflow of interface construction, light transport, and geometric refinement.

\input{sections/03_gaussian_interfaces_en}
\input{sections/03_coherent_refinement_en}
\input{sections/03_gaussian_guided_intersection_en}
\par
\input{figures/layer_budget}
\input{figures/accuracy_cost}
\input{tables/ablation_studies}
\input{tables/baseline_comparison}
\input{tables/image_formation}

%% file: figures/mls_mechanism.tex
\begin{figure}[t]
  \centering
  \includegraphics[width=\columnwidth]{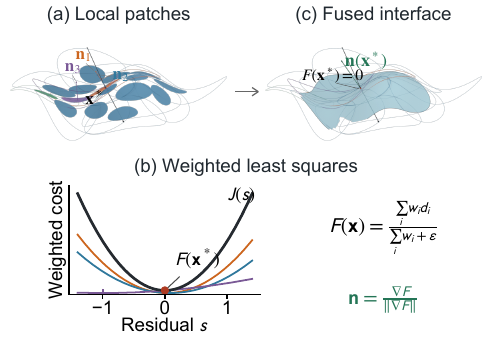}
  \caption{Gaussian-Field Hybrid Surfaces. (a) Irregular local patches and their normals. (b) Position-dependent weighted least-squares interpretation of the field value. (c) The fused zero-level surface and its gradient normal. Colors identify local patches, not separate physical interfaces.}
  \label{fig:mls-mechanism}
\end{figure}

%% file: sections/03_gaussian_interfaces_en.tex
\subsection{Gaussian-Field Hybrid Surfaces}
\label{sec:gaussian_interfaces}
Light transmission across multiple interfaces requires accurate local geometry and normal vectors, as each interaction alters the direction of subsequent light rays. We represent curvature using local polynomial surface patches and fuse them into intersectable optical interfaces via Gaussian weights, eliminating the need to explicitly store the connection relationships between surface patches. The degree of the polynomial controls the geometric expressiveness of each element, allowing the trade-off between accuracy and storage requirements to be evaluated independently for a fixed number of elements (Figure~\ref{fig:mls-mechanism}).

For element \(i\), the oriented input mesh provides the center \(\mathbf c_i\) and the unit normal \(\mathbf n_i^0\). Two orthonormal tangent vectors \(\mathbf t_{i,u}\) and \(\mathbf t_{i,v}\), together with the normal, form the local coordinate system \(\mathbf R_i=[\mathbf t_{i,u},\mathbf t_{i,v},\mathbf n_i^0]\). The query position \(\mathbf x\) in world coordinates is transformed into tangent coordinates \(u_i,v_i\) and normal height \(z_i\):
\begin{equation}
\bigl(u_i(\mathbf x),v_i(\mathbf x),z_i(\mathbf x)\bigr)^\top
=\mathbf R_i^\top(\mathbf x-\mathbf c_i).
\label{eq:gt-local-coordinates}
\end{equation}
Let \(N\) be the maximum total degree, \(\ell,m\) be non-negative integer exponents, and \(a_{i,\ell m}\) be coefficients. With the center and base tangent plane fixed, constant and linear terms are omitted; the local surface is defined by the zeros of the following residual:
\begin{equation}
d_i(\mathbf x)=z_i(\mathbf x)+
\sum_{\substack{\ell,m\geq0\\2\leq\ell+m\leq N}}
a_{i,\ell m}u_i(\mathbf x)^\ell v_i(\mathbf x)^m.
\label{eq:gt-local-residual}
\end{equation}

For \(N\geq1\), the coefficient count is \(M_N=\sum_{q=2}^{N}(q+1)\), equivalently \(M_N=(N+1)(N+2)/2-3\).
\(N=1\) corresponds to a plane; the default value \(N=3\) includes three quadratic terms and four cubic terms. Increasing the degree enhances local expressiveness at the cost of additional storage and evaluation overhead.

Given neighboring input samples \(\mathbf y_{ih}\) and fixed non-negative fitting weights \(\omega_{ih}\), the coefficient vector \(\mathbf a_i\) is obtained from
\begin{equation}
\mathbf a_i^*=\arg\min_{\mathbf a_i}
\sum_h\omega_{ih}\bigl[d_i(\mathbf y_{ih};\mathbf a_i)\bigr]^2.
\label{eq:gt-local-fit}
\end{equation}
We fit local polynomial coefficients using weighted least squares~\cite{alexa2001pointset,adamson2003approx}. Our implementation normalizes coordinates by a length scale, converts the fitted coefficients back to physical units, and checks rank and conditioning. The tuple \(\boldsymbol\gamma_i=(\mathbf c_i,\mathbf R_i,\mathbf a_i)\) consists of the center, local coordinate system, and surface coefficients, which together describe the local geometry and are the objects of the collaborative update described in Section~\ref{sec:coherent-refinement}.

The tangential scales \(s_{i,u},s_{i,v}\), the residual bandwidth \(\sigma_{i,n}\), and the smooth support window \(\psi_i\) define the influence range of the element:
\begin{equation}
w_i=\alpha_i\psi_i
\exp\!\left[-\frac12\left(
\frac{u_i^2}{s_{i,u}^2}
+\frac{v_i^2}{s_{i,v}^2}
+\frac{d_i^2}{\sigma_{i,n}^2}
\right)\right].
\label{eq:gt-query-weight}
\end{equation}
Here, all terms are evaluated at \(\mathbf x\). The amplitude \(\alpha_i\) is fixed, while the scales and support range follow the construction settings. Unlike the fitting weights \(\omega_{ih}\), the query weights \(w_i\) vary with the query location and geometry. The weight kernel is Gaussian in the residual coordinates \((u_i,v_i,d_i)\), so the influence range follows the local surface curvature rather than a general world-coordinate ellipsoid.

Inspired by weighted implicit surface constructions~\cite{ohtake2003mpu,shen2004polygonsoup}, we define the support measure \(C_k\) and interface field \(F_k\) for primitives \(\mathcal G_k\) assigned to interface \(k\) as
\begin{equation}
\begin{aligned}
C_k(\mathbf x)&=\sum_{i\in\mathcal G_k}w_i(\mathbf x),\\
F_k(\mathbf x)&=\frac{\sum_{i\in\mathcal G_k}w_i(\mathbf x)d_i(\mathbf x)}{C_k(\mathbf x)+\varepsilon}.
\end{aligned}
\label{eq:gt-blended-field}
\end{equation}
Here, \(\varepsilon>0\) stabilizes the denominator. At a fixed query location, \(F_k(\mathbf x)\) is the minimizer of the following objective with respect to the scalar \(\zeta\):
\[
F_k(\mathbf x)=\underset{\zeta\in\mathbb R}{\arg\min}
\left[\sum_{i\in\mathcal G_k}w_i(\mathbf x)
\bigl(\zeta-d_i(\mathbf x)\bigr)^2+\varepsilon\zeta^2\right].
\]
This interpretation requires no additional iterative solution. With all terms evaluated at \(\mathbf x\), the gradient of the same field is
\begin{equation}
\nabla F_k=
\frac{\sum_{i\in\mathcal G_k}\left[
w_i\nabla d_i
+\bigl(d_i-F_k\bigr)\nabla w_i
\right]}{C_k+\varepsilon}.
\label{eq:gt-field-gradient}
\end{equation}
Given a support threshold \(\tau_C\), the optical interface and its normal are defined as
\begin{equation}
\begin{aligned}
\mathcal S_k&=\{\mathbf x:F_k(\mathbf x)=0,\ C_k(\mathbf x)\geq\tau_C\},\\
\mathbf n_k(\mathbf x)&=\frac{\nabla F_k(\mathbf x)}{\|\nabla F_k(\mathbf x)\|}.
\end{aligned}
\label{eq:gt-interface-normal}
\end{equation}
Normals are defined at roots that are differentiable, valid, and have nonzero gradients. Both the residual-dependent weight-kernel derivative and the support-window derivative are retained; averaging local surface normals generally yields a different direction. Therefore, intersection and refraction use the same surface, and each interface retains the medium affiliation on both sides.

%% file: sections/03_coherent_refinement_en.tex
\subsection{Transport-Coupled Refinement}
\label{sec:coherent-refinement}
An update to one interface alters the light rays directed toward subsequent interfaces; therefore, the corrections for each layer, calculated independently, may no longer be consistent when combined. We evaluate each candidate update by tracing the complete primary light path, share parameters across multiple viewpoints, and utilize the currently accepted geometric states of other shells.

For \(S\) shells, the parameter vector \(\boldsymbol\theta=(\theta_1,\ldots,\theta_S)\) contains one local deformation control variable per shell, shared by the inner and outer interfaces of that shell. Let \(s(i)\) denote the shell to which element \(i\) belongs, and \(\mathcal M^0\) denote the coarse input mesh. Cooperative refitting updates the local geometry defined in Section~\ref{sec:gaussian_interfaces}:
\begin{equation}
\begin{aligned}
\boldsymbol\gamma_i(\boldsymbol\theta)
&=\mathcal U_i(\mathcal M^0,\theta_{s(i)})\\
&=\bigl(\mathbf c_i(\boldsymbol\theta),\mathbf R_i(\boldsymbol\theta),\mathbf a_i(\boldsymbol\theta)\bigr).
\end{aligned}
\label{eq:gt-coherent-update}
\end{equation}
The update operator \(\mathcal U_i\) causes the fixed mesh samples to move with the deformation, reconstructs the local coordinate system, and re-solves the coefficients according to Equation~\eqref{eq:gt-local-fit}. Substituting the updated results into Equations~\eqref{eq:gt-local-coordinates}--\eqref{eq:gt-interface-normal} yields the \(\boldsymbol\theta\)-dependent fusion field \(F_k(\mathbf x;\boldsymbol\theta)\), support \(C_k(\mathbf x;\boldsymbol\theta)\), and normal vector \(\mathbf n_k(\mathbf x;\boldsymbol\theta)\).

Observation \(q\) originates from a ray emitted by a calibrated camera. At the \(j\)th interaction, the current ray segment is \(\mathbf r_{q,j}(t)=\mathbf o_{q,j}+t\mathbf v_{q,j}\), with \(\|\mathbf v_{q,j}\|_2=1\), where \(\mathbf o_{q,j}\) is the segment origin, \(\mathbf v_{q,j}\) is the unit direction, and \(\mu_{q,j}\) is the incident medium. The query operator \(\mathcal Q\) returns the accepted interface and ray parameter; Section~\ref{sec:gaussian-guided-intersection} describes an optional element-guided implementation. Propagation follows Snell's law~\cite{pharr2023pbrt} through the recursive relationship
\begin{equation}
\begin{aligned}
(k_{q,j},t_{q,j})&=\mathcal Q(\mathbf r_{q,j},\mu_{q,j};\boldsymbol\theta),\\
\mathbf x_{q,j}&=\mathbf r_{q,j}(t_{q,j}),\\
\mathbf v_{q,j+1}&=\operatorname{Snell}(\mathbf v_{q,j},\mathbf n_{q,j},\rho_{q,j}),
\end{aligned}
\label{eq:gt-path-recurrence}
\end{equation}
where the current interaction normal is
\[
\mathbf n_{q,j}
=\mathbf n_{k_{q,j}}(\mathbf x_{q,j};\boldsymbol\theta)
=\frac{\nabla_{\mathbf x}F_{k_{q,j}}(\mathbf x_{q,j};\boldsymbol\theta)}{\|\nabla_{\mathbf x}F_{k_{q,j}}(\mathbf x_{q,j};\boldsymbol\theta)\|},
\]
and \(\rho_{q,j}\) is the ratio of the refractive indices of the incident and transmitted media. The next ray segment originates from the current interaction point, utilizing the transmitted medium and existing self-intersection exclusion rules. All states depend on \(\boldsymbol\theta\); this recursive process propagates each geometric change to subsequent intersection points and the final background prediction \(\widehat{\mathbf b}_q(\boldsymbol\theta)\).

A fixed training set \(\mathcal D\) provides multi-view background coordinates \(\mathbf b_q\), a form of refractive correspondence supervision~\cite{lyu2020drt,li2023neto}. When corresponding observations are available, rays passing through a single shell and those passing through multiple shells constrain different interface combinations. All observations evaluate the same parameter vector. Invalid light paths are penalized with a fixed \(\kappa\):
\begin{equation}
\ell_q(\boldsymbol\theta)=
\begin{cases}
\|\widehat{\mathbf b}_q(\boldsymbol\theta)-\mathbf b_q\|_2^2,&\text{valid path},\\
\kappa,&\text{otherwise}.
\end{cases}
\label{eq:gt-path-loss}
\end{equation}
Valid light paths must complete the specified interface sequence and reach the background in the positive direction. The joint objective is minimized within the predefined domain \(\Theta\):
\begin{equation}
\mathcal L(\boldsymbol\theta)=
\frac{1}{|\mathcal D|}\sum_{q\in\mathcal D}\ell_q(\boldsymbol\theta)
+\lambda\|\boldsymbol\theta\|_2^2,\qquad\boldsymbol\theta\in\Theta.
\label{eq:gt-joint-objective}
\end{equation}
Here, \(\lambda\) is a fixed penalty weight, and the denominator counts all observations, including those with failed light paths. During bounded coordinate pattern search~\cite{kolda2003directsearch}, each candidate is evaluated with the other shells at their currently accepted controls. This interlinks candidate selection across different shells, while \(\mathcal U_i\) synchronously updates the center, coordinate system, and surface coefficients of each element. Within a fixed search budget, parameters are selected based solely on training data; evaluation is then performed using held-out optical paths and per-interface geometry. Improved background landing accuracy does not imply that the geometry has been uniquely reconstructed. Complete Fresnel imaging is evaluated separately; see Supplementary Section~S4 for implementation details.

%% file: sections/03_gaussian_guided_intersection_en.tex
\subsection{Primitive-Guided Implicit Intersection}
\label{sec:gaussian-guided-intersection}
\label{sec:multi-interface-transport}
Equation~\eqref{eq:gt-path-recurrence} uses an optical path recursion to repeatedly query the interface field constructed in Section~\ref{sec:gaussian_interfaces}. Although individual local residuals are polynomials, the position-dependent weights typically cause the fused field along the ray to no longer be a polynomial. Following local-fit guidance in point-set ray tracing~\cite{adamson2003ray,adamson2003approx}, this module reuses the primitive's support range and local roots to increase sampling density near potential intersection points.

For the current ray segment, abbreviate \(\mathbf r_{q,j}\) as \(\mathbf r(t)=\mathbf o+t\mathbf v\), and omit the fixed \(\boldsymbol\theta\). Substituting Equation~\eqref{eq:gt-local-coordinates} into Equation~\eqref{eq:gt-local-residual} yields
\begin{equation}
P_i(t)=d_i(\mathbf r(t)),\qquad\deg P_i\leq N.
\label{eq:gt-ray-polynomial}
\end{equation}
The search interval is clipped using the bounding rectangle of the original tangential support range to obtain \(I_i\). The numerical real roots, stationary points, and residual-support boundary candidates of \(P_i\) within the interval are used to guide sampling. Fixed samples are added within the interval and near the candidates, and merged with the original coarse samples:
\begin{equation}
\mathcal T=\mathcal T_{\mathrm{coarse}}\cup\mathcal T_{\mathrm{guide}}.
\label{eq:gt-guided-samples}
\end{equation}
Here, \(\mathcal T_{\mathrm{coarse}}\) and \(\mathcal T_{\mathrm{guide}}\) contain the original and additional ray-parameter samples, respectively. The implementation has been evaluated using \(N=3\).

At the merged sampling locations \(t\in\mathcal T\), we calculate the unchanged fusion field \(F_k(\mathbf r(t))\) and support measure \(C_k(\mathbf r(t))\) from Equation~\eqref{eq:gt-blended-field}, and perform bisection on sign-changing intervals with valid support. For candidate intersection points \(\mathbf x^*=\mathbf r(t^*)\), the following conditions must be satisfied:
\begin{equation}
\begin{aligned}
|F_k(\mathbf x^*)|\leq\tau_F,\qquad&C_k(\mathbf x^*)\geq\tau_C,\\
&\|\nabla F_k(\mathbf x^*)\|_2\geq\tau_g,
\end{aligned}
\label{eq:gt-root-acceptance}
\end{equation}
where \(\tau_F,\tau_C,\tau_g\) are fixed thresholds for the residual, support, and gradient, respectively. Let \(\mathcal A(\mathbf r,\mu;\boldsymbol\theta)\) collect the interface/ray-parameter pairs \((k,t)\) that have passed validation and are medium-compatible, after applying the original root-merging rule and self-intersection exclusion \(t>t_{\min}\). The query operator used in Equation~\eqref{eq:gt-path-recurrence} is
\begin{equation}
\mathcal Q(\mathbf r,\mu;\boldsymbol\theta)
=\underset{(k,t)\in\mathcal A(\mathbf r,\mu;\boldsymbol\theta)}{\arg\min}\ t.
\label{eq:gt-query-operator}
\end{equation}
When the candidate set is empty, the existing no-hit or failure handling is applied; the normal vector is derived from the field gradient defined in Equation~\eqref{eq:gt-interface-normal}.

During imaging, the same valid interactions extend the primary-transmission recursion to Fresnel-weighted reflection and transmission~\cite{pharr2023pbrt}. The contribution of resolved paths is
\[
\mathbf C_{\mathrm{res}}=
\sum_{\pi\in\mathcal P_{\mathrm{res}}}
\boldsymbol\beta_\pi\odot\mathbf L_\pi,
\]
where \(\mathcal P_{\mathrm{res}}\) is the set of resolved paths, \(\mathbf L_\pi\) is the terminal radiance, and \(\boldsymbol\beta_\pi\) is the cumulative path throughput incorporating Fresnel factors, refractive-index-dependent radiance scaling, and specified attenuation. The operator \(\odot\) denotes componentwise multiplication. Event and node budgets limit the tracing overhead. When the source radiance is bounded and the path has no hard failures, the weights of the remaining branches provide bounds on the unresolved contribution.

%% file: figures/layer_budget.tex
\begin{figure*}[t]
\centering
\includegraphics[width=\textwidth]{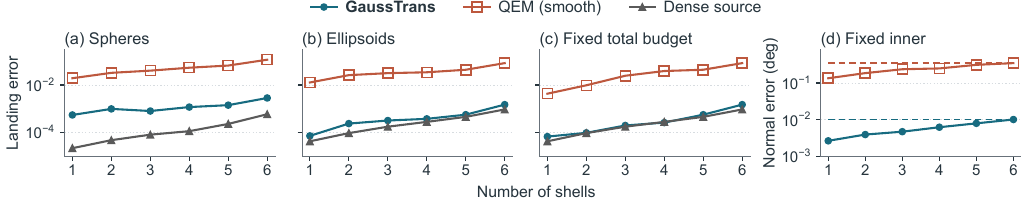}
\caption{Layer count and budget. (a, b) Spherical/ellipsoidal shells, fixed per-interface capacity. (c) Ellipsoids, fixed total storage. Landing errors use 65 fixed primary rays/five views; dense source has higher storage. (d) With the inner surface held fixed, solid lines evenly distribute a fixed total storage budget across interfaces; dashed lines keep the per-interface representation capacity fixed. Units: scene units/degrees.}
\label{fig:layer_budget}
\end{figure*}

%% file: figures/accuracy_cost.tex
\begin{figure*}[t]
  \centering
  \includegraphics[width=\textwidth]{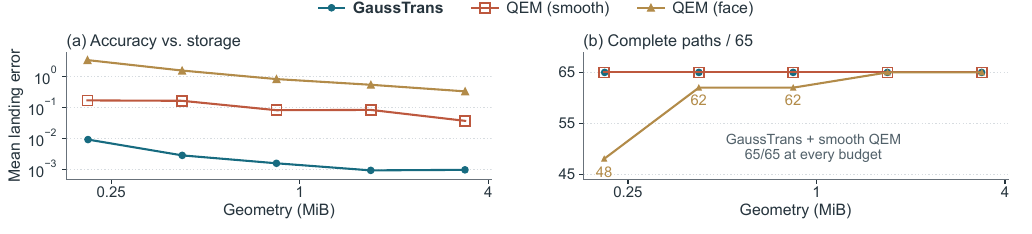}
  \caption{Six-shell accuracy and path completion under five approximately byte-matched Gaussian/QEM~\cite{garland1997quadrics} budgets. (a) Mean landing error on the same 44 rays completed by all 15 configurations. (b) Complete 24-event paths/65; GaussTrans and smooth QEM both reach 65 at every budget. Storage: uncompressed numerical geometry. Errors: scene units.}
  \label{fig:accuracy_cost}
\end{figure*}

%% file: tables/ablation_studies.tex
\begin{table*}[t]
\centering
\caption{Local-degree and update ablations. Errors use scene units/degrees at the indicated scales; bold marks best means.}
\label{tab:ablations}
\small
\setlength{\tabcolsep}{1.3pt}
\renewcommand{\arraystretch}{1.08}
\begin{minipage}[t]{0.515\textwidth}
\vspace{0pt}
\centering
\begin{tabular*}{\linewidth}{@{\extracolsep{\fill}}lrrrrrrr@{}}
\toprule
\multicolumn{8}{@{}l}{(a) Local surface degree}\\
Degree $N$ & 2 & 3 & 4 & 5 & 6 & 7 & 8 \\
\midrule
Shape coeffs. & 3 & 7 & 12 & 18 & 25 & 33 & 42 \\
Landing ($10^{-3}$) & 2.564 & 1.498 & 0.851 & 0.616 & 0.637 & \textbf{0.488} & 0.507 \\
Angle ($10^{-2}$ deg) & 4.189 & 2.428 & 1.394 & 1.011 & 1.045 & \textbf{0.794} & 0.825 \\
Time (s) & 8.149 & 8.046 & 9.706 & 11.144 & 13.036 & 14.913 & 17.078 \\
Bank (KiB) & 712.5 & 862.5 & 1050.0 & 1275.0 & 1537.5 & 1837.5 & 2175.0 \\
\bottomrule
\end{tabular*}
\par\vspace{3pt}
\begin{minipage}{\linewidth}
\small
Six shells, 400 Gaussians per interface; default degree $N=3$. Gaussian parameter-bank storage includes redundant local frames and metadata.
\end{minipage}
\end{minipage}\hfill
\begin{minipage}[t]{0.465\textwidth}
\vspace{0pt}
\centering
\begin{tabular*}{\linewidth}{@{\extracolsep{\fill}}llrrrr@{}}
\toprule
\multicolumn{6}{@{}l}{(b) Within-primitive updates} \\
Scene & Update & Landing $\downarrow$ & Position $\downarrow$ & Normal $\downarrow$ & Agree. $\uparrow$ \\
\midrule
Sphere & Center-only & 0.007381 & 0.008950 & 0.823 & 511/512 \\
Sphere & Coherent & \textbf{0.002550} & \textbf{0.002926} & \textbf{0.226} & \textbf{512/512} \\
Ellipsoid & Center-only & 0.007908 & 0.008897 & 0.818 & 509/512 \\
Ellipsoid & Coherent & \textbf{0.002967} & \textbf{0.001271} & \textbf{0.164} & \textbf{512/512} \\
\bottomrule
\end{tabular*}
\par\vspace{3pt}
\begingroup
\setlength{\aboverulesep}{1.5pt}
\setlength{\belowrulesep}{1.5pt}
\begin{tabular*}{\linewidth}{@{\extracolsep{\fill}}lrrr@{}}
\toprule
\multicolumn{4}{@{}l}{(c) Cross-shell context: landing ($10^{-3}$)} \\
Input & Initial & Independent & Full \\
\midrule
A & 11.884 & 11.498 & \textbf{8.352} \\
B & 12.015 & 12.527 & \textbf{8.015} \\
\bottomrule
\end{tabular*}
\endgroup
\end{minipage}

\par\vspace{3pt}
\begin{minipage}{\textwidth}
\small
(b) Both double-shell arms use 16 observations, two controls, and 13 evaluations. Landing: four held-out rays/scene; position/normal: 24 probes; agreement: all 512 test rays including background, TIR states, and event prefixes, not 512 transmitted paths. Center-only selects zero deformation in both scenes.
(c) A/B: alternating/reversed input perturbations. All variants complete 39/39 paths per input; the previously examined evaluation bank is excluded from selection. Search times: independent 47.2/47.9\,s; full 48.2/48.0\,s.
\end{minipage}
\par\vspace{8pt}
\end{table*}

%% file: tables/baseline_comparison.tex
\begin{table*}[t]
\centering
\caption{RefRef sphere: 100 test views at $512\times512$; metrics are averaged over views. Only ours and Mesh use the same inputs. KiB = $2^{10}$ bytes, MB = $10^6$ bytes; primitive counts follow different accounting conventions (Supplementary Sec.~S1).}
\label{tab:baselines}
\small
\setlength{\tabcolsep}{4pt}
\renewcommand{\arraystretch}{1.06}
\begin{tabular*}{\textwidth}{@{\extracolsep{\fill}}lrrrrl@{}}
\toprule
Method & PSNR $\uparrow$ & SSIM $\uparrow$ & LPIPS $\downarrow$ & Object MAE $\downarrow$ & Representation / asset scope \\
\midrule
GaussTrans & 23.49 & 0.9085 & 0.0665 & 0.1409 & 82.85 KiB (geometry) \\
Mesh (same inputs) & 23.45 & 0.9068 & 0.0838 & 0.1414 & 2140.17 KiB (geometry) \\
RT-Splatting~\cite{shi2026rtsplatting} & 23.44 & 0.9010 & 0.0877 & 0.1442 & 125.77 MB (inference) \\
TransparentGS~\cite{huang2025transparentgs}$^{a}$ & 22.16 & 0.8923 & 0.0953 & 0.1480 & 1.586M Gaussians \\
TNSR~\cite{deng2024tnsr} & 22.42 & 0.8741 & 0.1600 & 0.1235 & 67.78 MB (inference) \\
R3F~\cite{yin2025refref} (estimated mesh) & 24.54 & 0.8999 & 0.0925 & 0.1263 & radiance field + mesh \\
3DGS~\cite{kerbl2023gaussians} & 23.61 & 0.9025 & 0.0876 & 0.1439 & radiance Gaussians \\
2DGS~\cite{huang2024surfsplatting} & 23.64 & 0.9042 & 0.0908 & 0.1420 & radiance Gaussians \\
\bottomrule
\end{tabular*}

\par\vspace{3pt}
\begin{minipage}{\textwidth}
\small
$^{a}$ TransparentGS learns IOR (1.46854); ours/Mesh and TNSR fix it at 1.5.
\end{minipage}
\end{table*}

%% file: tables/image_formation.tex
\begin{table*}[t]
\centering
\caption{RefRef torus~\cite{yin2025refref}; QEM~\cite{garland1997quadrics}: one $512\times512$ view, four-crossing region. Brackets: RGB MAE bounds from unresolved transport, not confidence intervals. All hard failures remain counted.}
\label{tab:image_formation}
\small
\setlength{\tabcolsep}{4pt}
\renewcommand{\arraystretch}{1.05}
\begin{tabular*}{\textwidth}{@{\extracolsep{\fill}}lrrrrr@{}}
\toprule
Method & \shortstack{Core geometry\\(KiB) $\downarrow$} & \shortstack{Normal cache\\(KiB)} & \shortstack{Common-pixel\\MAE bounds $\downarrow$} & \shortstack{Hard pixels\\all $\downarrow$} & \shortstack{Hard pixels\\four-crossing $\downarrow$}\\
\midrule
GaussTrans & 96.88 & 0.00 & [0.07490, 0.07534] & 0 & 0 \\
QEM matched (F) & 96.82 & 0.00 & [0.17436, 0.17541] & 0 & 0 \\
QEM matched (S) & 96.82 & 32.27 & [0.08844, 0.08896] & 768 & 137 \\
QEM double (S) & 193.71 & 64.57 & [0.06980, 0.07017] & 382 & 69 \\
Dense Mesh (F) & 648.00 & 0.00 & [0.11830, 0.11879] & 0 & 0 \\
Dense Mesh (S) & 648.00 & 216.00 & [0.04808, 0.04845] & 166 & 28 \\
\bottomrule
\end{tabular*}

\par\vspace{3pt}
\begin{minipage}{\textwidth}
\small
F/S: face/smooth normals. MAE: the same 5,560 hard-failure-free pixels across nine archived configurations. Failure denominators: 262,144 (all), 6,069 (four-crossing); the 509 excluded pixels remain counted. Core bytes exclude listed normal caches and other runtime data. Four crossings do not imply four nested objects. Bounds cover frozen transport, not omitted appearance effects.
\end{minipage}
\end{table*}

%% file: sections/04_experiments.tex
\section{Experiments}
\label{sec:experiments}
\subsection{Experimental Setup}
We evaluate interface representation, geometric refinement, and imaging on spherical shells, ellipsoidal shells, quartic surface shells, and the RefRef~\cite{yin2025refref} scene. Cubic surface patches are used by default; the newly added refinement and guided intersection tests cover 16 input sets and 1,024 pixels, respectively. Same-input controls share geometry, cameras, media, and tracing budgets; capacity experiments are matched based on geometric storage requirements. We report geometric and optical path errors, RGB error ranges, PSNR, SSIM~\cite{wang2004ssim}, and VGG-LPIPS~\cite{zhang2018lpips}, while also quantifying storage and runtime. Refinement parameters were selected solely from training observations; evaluation used held-out viewpoints. Other baselines retained their original priors; timing comparisons were conducted under identical execution conditions.

\subsection{Surface Construction and Accuracy}
\label{sec:accuracy_cost}
We verify its accuracy in continuous multi-interface propagation. First, we construct different surface representations from the same input geometry. Since errors in preceding intersection points and normals can affect subsequent propagation, we evaluate the entire light path rather than just local surface fits. Figure~\ref{fig:layer_budget} uses 65 fixed rays from five viewpoints, covering one to six layers of spherical and ellipsoidal shells. With a setting of 400 cubic Gaussian functions per interface, both GaussTrans and the QEM mesh~\cite{garland1997quadrics} stored with the same geometry successfully traced all specified paths, with GaussTrans exhibiting a lower average background landing point error. When the total capacity is fixed, increasing the number of interfaces raises local normal errors; maintaining the capacity per interface preserves local accuracy. Therefore, multi-layer evaluation requires simultaneous consideration of both continuous propagation and the representation capacity of each interface.

We further validated these results on non-convex geometries and final images. On the RefRef ring~\cite{yin2025refref}, under comparable geometric storage, GaussTrans achieved lower RGB error thresholds for total effective pixels, with fewer hard failures across the entire image compared to Smooth QEM; all hard failures were statistically preserved (Table~\ref{tab:image_formation}, Supplementary Figure~S2). Across 100 test viewpoints of the glass sphere, Table~\ref{tab:baselines} compares TransparentGS~\cite{huang2025transparentgs}, TNSR~\cite{deng2024tnsr}, R3F~\cite{yin2025refref}, RT-Splatting~\cite{shi2026rtsplatting}, 3DGS~\cite{kerbl2023gaussians}, and 2DGS~\cite{huang2024surfsplatting}. GaussTrans achieved the highest SSIM (0.9085) and lowest LPIPS (0.0665) in the table, but R3F had a higher PSNR, while TNSR and R3F had lower MAE in object regions. This set of comparisons evaluates imaging quality, while the ability to propagate continuously across multiple interfaces is verified by the shell-and-ring experiments.

Memory compression is another benefit beyond the aforementioned accuracy. On a six-layer rotating eccentric ellipsoid, cubic Gaussian, quadratic B\'ezier, and cubic B\'ezier methods~\cite{farin2002cagd}---each requiring approximately 861--863 KiB of geometric memory---all completed 65 paths, with average landing point errors of 0.001498, 0.003337, and 0.005653, respectively. This indicates that GaussTrans still maintains an advantage in average accuracy compared to these two surface patch constructions. In a five-level comparison of approximate storage matching with 100 to 1,600 Gaussian elements per interface, both GaussTrans and Smooth QEM completed 65 out of 65 paths, with the former reducing the average landing point error by 95.1\% to 98.9\% (Figure~\ref{fig:accuracy_cost}). The error curves in the figure are based on 44 rays processed using 15 different configurations, while the method averages are calculated using all 65 rays. Comparisons using the same input mesh for the glass sphere also show that GaussTrans achieves similar PSNR and lower LPIPS with approximately 1/25.8 of the geometric storage. External methods retain their respective priors. The current advantages primarily lie in the trade-off between accuracy and geometric storage: denser meshes can still achieve greater accuracy, and Gaussian tracking takes significantly longer than smooth QEM mesh in the five-level capacity experiments.

\subsection{Ablation Studies}
\label{sec:ablations}
\emph{Order of the surface.}\nopagebreak[4] Under fixed centers, local frames, and support domains and with 400 Gaussian primitives per interface, surfaces of the second through eighth orders all completed 65 out of 65 light paths (Table~\ref{tab:ablations}(a)); the linear surface completed 61 out of 65, with the remaining four paths undergoing total internal reflection. Compared to the second-order fit, the third-order fit reduced the average landing point error by 41.6\%, with improvements in the mean values across all five viewpoints and in 54 out of 65 light rays. Higher-order fits trade increased storage and runtime for accuracy, but the sixth- and eighth-order fits showed a decline in performance.

\emph{Coordinated updates within primitives.}\nopagebreak[4] By simultaneously updating the local coordinate system and surface coefficients with matched center displacements and optimization budgets, improvements were achieved in the background landing-point errors in held-out views, as well as in geometric and light path consistency for both double-shell scenarios (Table~\ref{tab:ablations}(b)), although isolated cases of deterioration still occurred.

\emph{Cross-shell updates.}\nopagebreak[4] The six-layer ellipsoidal shell uses six control variables and ten synthetic background correspondences. Independent corrections were selected separately with the other shells at their initial geometries and then combined; full updates utilized corrections already accepted by other shells. Both approaches employed collaborative refitting, 25 search evaluations, and a single final training check. Compared to independent corrections, the full update reduced the average landing point error across unseen viewpoints by 27.4\% and 36.0\% on the two input sets, respectively, with improvements observed for all 39 paired rays in each set (Table~\ref{tab:ablations}(c)). Compared to the initialization, the mean errors decreased by 29.7\% and 33.3\%, respectively; updates involving only center shifting and smooth-normal QEM updates resulted in deterioration.

\emph{Primitive-guided intersection finding.}\nopagebreak[4] Under fixed fusion field and intersection acceptance criteria, we compared results with and without guidance. Among the 8 existing hard-failure pixels in the six-shell scene, enabling guidance reduced the number of hard failures from 8 to 2. In 41 grazing-ray perturbations, for the 33 cases with existing observed intersection points, the number of recovered points increased from 20 to 33. These two sets of directed diagnostics support the restorative effect of guidance on difficult queries, though it comes with additional computational overhead. Appendix~S7 reports per-input and per-interface errors, failures, and timings for the cross-shell ablation.

\subsection{Complete Imaging Across Multiple Interfaces}
\label{sec:image_formation}
Beyond evaluating the main transmission path, we further traced the Fresnel reflection and transmission branches~\cite{pharr2023pbrt} to verify whether the Gaussian interface can support complete image generation. Figure S1 in the Supplementary Materials shows 512$\times$512 rendering results for the six-shell scene, including an analytical reference, GaussTrans, storage-matched QEM with smooth normals, and a dense source mesh, all using the same light transport budget. This experiment extends the evaluation from a single main path to multi-branch color accumulation; however, the limited budget still leaves a reflection tail: the average unresolved weight of the foreground for the analytical reference, GaussTrans, and dense source mesh is approximately 5.3\%. The resulting images therefore retain unresolved contribution intervals rather than representing converged transport.

\subsection{Limitations}
The accuracy of the GaussTrans interface depends on the quality of the input geometry and local coverage. For thin-walled structures, adjacent surfaces, or regions with rapidly varying curvature, local surface fitting and Gaussian blending may still result in positional and normal deviations, or even introduce extra intersection points or omit valid ones. These local errors may be amplified during continuous refraction, limiting the applicability of the current method to complex transparent structures.

Primitive-Guided Implicit Intersection uses local polynomial roots to propose candidate positions; however, these roots do not necessarily correspond to the zeros of the fusion field and still require verification using actual field values and support conditions. Nevertheless, the generation and verification of candidates also increase computational overhead. Therefore, this module is currently used to assist with certain difficult queries.

%% file: sections/05_conclusion.tex
\GaussTransCompactSection{Conclusion}
\label{sec:conclusion}
We presented GaussTrans, compact Gaussian--MLS interfaces for high-order transport with same-field intersections and normals. Given coarse geometry and known media, Transport-Coupled Refinement reduces held-out path error on perturbed inputs. At comparable geometry storage, one-to-six-shell experiments reduce primary-transmission errors versus evaluated meshes. On the nonconvex RefRef torus, GaussTrans achieves lower RGB MAE bounds and fewer hard failures than smooth QEM at comparable core storage. These results characterize the accuracy--storage--runtime trade-off.

%% file: supplement_body.tex
\clearpage
\onecolumn
\begingroup
\GaussTransBodyAlignmentfalse\normalsize
\nonfrenchspacing
\makeatletter
\patchcmd{\cvprsection}{-8pt}{-10pt}{}{\PackageError{GaussTrans}{Supplement heading before-gap failed}{}}
\patchcmd{\cvprsection}{5pt}{7pt}{}{\PackageError{GaussTrans}{Supplement heading after-gap failed}{}}
\let\@addtocurcol\GaussTransKernelAddToCurCol
\makeatother
\input{supplement/sections/layout_style}
\setcounter{section}{0}
\setcounter{subsection}{0}
\setcounter{figure}{0}
\setcounter{table}{0}
\setcounter{equation}{0}
\renewcommand{\thesection}{\Alph{section}}
\renewcommand{\thesubsection}{\thesection.\arabic{subsection}}
\renewcommand{\thetable}{S\arabic{table}}
\renewcommand{\thefigure}{S\arabic{figure}}
\renewcommand{\theequation}{S\arabic{equation}}
\renewcommand{\theHsection}{supp.\arabic{section}}
\renewcommand{\theHsubsection}{supp.\arabic{section}.\arabic{subsection}}
\renewcommand{\theHsubsubsection}{supp.\arabic{section}.\arabic{subsection}.\arabic{subsubsection}}
\renewcommand{\theHtable}{supp.\arabic{table}}
\renewcommand{\theHfigure}{supp.\arabic{figure}}
\renewcommand{\theHequation}{supp.\arabic{equation}}
\twocolumn[\begin{center}
\vspace*{12pt}
{\Large\bfseries GaussTrans: A Gaussian Interface Framework\\
for Multi-Layer Refractive Transport\par}
\vspace{8pt}
{\large Supplementary Material\par}
\vspace{14pt}
\end{center}]
\pdfbookmark[0]{Supplementary Material}{supplementary-material}
\input{supplement/sections/navigation}
\input{supplement/sections/a_protocols}
\FloatBarrier
\input{supplement/sections/b_results}
\input{supplement/sections/c_ablations}
\par\penalty-10000
\clearpage
\endgroup

%% file: supplement/sections/layout_style.tex
\setlength{\textfloatsep}{8pt plus 2pt minus 1pt}
\setlength{\dbltextfloatsep}{8pt plus 2pt minus 1pt}
\setlength{\intextsep}{8pt plus 2pt minus 1pt}
\setlength{\floatsep}{8pt plus 2pt minus 1pt}
\setlength{\dblfloatsep}{8pt plus 2pt minus 1pt}
\captionsetup{skip=5pt,singlelinecheck=false,justification=raggedright}
\setlength{\parskip}{0pt plus 0.3pt}
\flushbottom
\makeatletter
\setlength{\@dblfptop}{0pt}
\pretocmd{\cvprsection}{\Needspace{5\baselineskip}}{}{}
\pretocmd{\cvprsubsection}{\Needspace{4\baselineskip}}{}{}
\makeatother
\widowpenalty=10000
\clubpenalty=10000
\displaywidowpenalty=10000

%% file: supplement/sections/navigation.tex
This supplementary material includes the experimental setup (A), full results (B), and ablation studies (C). The definitions of the methods remain consistent with those in the main text; this document provides only the specific implementation details required for each experiment. The correspondence between the main paper and these experiments is outlined in the table below. Tables S1--S9 and Figures S1--S2 retain their original numbering.
\par\vspace{6pt}\noindent\begingroup\centering
\small\setlength{\tabcolsep}{4pt}
\begin{tabularx}{\columnwidth}{@{}lX@{}}
\toprule
Main-paper item & Supplementary content \\
\midrule
Figure 1 & \ref{appendix:sec:supp-real-ball}: captured Ball and baseline settings \\
Figure 4 & \ref{appendix:sec:supp-input-comparison}: layer count and capacity \\
Figure 5 & \ref{appendix:sec:supp-capacity}: storage, accuracy, and tracing cost \\
Table 1(a,b) & \ref{appendix:sec:order-within}: polynomial order and coherent updating \\
Table 1(c) & \ref{appendix:sec:supp-additional-ablations}: cross-shell updates \\
Table 2 & \ref{appendix:sec:storage-accounting}, \ref{appendix:sec:supp-sphere}: sphere storage and transport \\
Table 3 & \ref{appendix:sec:supp-input-comparison}, \ref{appendix:sec:supp-qualitative}: torus settings and images \\
Sections 4.1, 4.3 & \ref{appendix:sec:supp-refinement16}--\ref{appendix:sec:supp-guided1024}: extended ablations \\
Section 4.4 & \ref{appendix:sec:supp-qualitative}: full Fresnel images \\
\bottomrule
\end{tabularx}
\par\endgroup\vspace{6pt}

%% file: supplement/sections/a_protocols.tex
\Needspace{10\baselineskip}
\section{Experimental Protocols and Comparison Conditions}
\label{appendix:sec:protocols}
\subsection{Shared conventions and measurement scope}
\label{appendix:sec:shared-protocol}
A hollow shell contains two interfaces; a light ray passing entirely through $L$ nested shells undergoes $4L$ interface interactions. Consequently, a six-shell configuration comprises 12 interfaces, corresponding to 24 distinct transmission events. In the composite shell experiments, the refractive indices for air and glass are set to 1.0 and 1.5, respectively. Analytical reference data are used to generate ground truth for observation and evaluation, while the Gaussian geometry is derived by fitting to the input meshes. Positional errors are expressed in the scene coordinate units, and angular errors are measured in degrees. PSNR, SSIM, and VGG-LPIPS metrics are calculated based on the floating-point image outputs, with the final values representing the average across all viewpoints.

Face-normal Mesh uses triangle normals; smooth-normal Mesh uses normalized barycentric interpolation of vertex normals with its original side-validity rules. Only declared same-input pairs share priors.

Intersection misses, valid total internal reflection (TIR) events, and truncations due to computational budget constraints are tracked separately. The upper and lower bounds for RGB values are used to define the contribution of light paths that remain unresolved under a given radiance model; pixels classified as ``hard failures'' are excluded from the valid prediction set. The ``RGB error threshold'' mentioned in Section~4.2 of the main text refers to the upper and lower bounds of the mean absolute error (MAE) calculated over the set of mutually valid pixels (as presented in Table~3), with failed pixels accounted for separately.

\paragraph{Execution conditions.}
The capacity, refinement, and guided-intersection experiments use CPU-based double-precision computation. Runtime comparisons use matched execution settings, with construction and tracing times reported separately.

\subsection{Storage accounting}
\label{appendix:sec:storage-accounting}
Table~\ref{appendix:tab:compactness} lists the storage requirements for each component in the glass sphere and torus experiments. Since different metrics are used for geometric representations, inference models, training states, and the number of primitives, these are reported separately. For the glass sphere, the total model storage is broken down into that required for geometry versus RGB inference; for the torus, the core geometry and the additional normal cache are listed separately.

\setcounter{table}{0}
\begin{table}[t]
\centering
\caption{Storage by component. KiB = $2^{10}$ bytes; MB = $10^6$ bytes. Geometry counts numerical arrays; inference measures stored models without optimizer state.}
\label{appendix:tab:compactness}
\small
\setlength{\tabcolsep}{3pt}
\begin{tabularx}{\columnwidth}{@{}>{\raggedright\arraybackslash}X>{\raggedright\arraybackslash}Xr@{}}
\toprule
Scene / method & Counted component & Size \\
\midrule
Sphere / ours & Interface state & 82.85 KiB \\
Sphere / Mesh & Vertices + faces & 2140.17 KiB \\
Torus / ours & Core geometry & 96.88 KiB \\
Torus / dense Mesh & Vertices + faces & 648.00 KiB \\
\addlinespace[2pt]
Sphere / RT-Splatting & RGB-inference model & 125.77 MB \\
Sphere / TNSR & RGB-inference model & 67.78 MB \\
Sphere / TNSR & Training state & 183.49 MB \\
Sphere / TransparentGS & FG + BG primitives & 1,585,532 \\
\bottomrule
\end{tabularx}
\par\vspace{3pt}
\begin{minipage}{\columnwidth}
\small
Geometry excludes shared backgrounds and runtime caches. Sphere interface state includes query domains and interface identifiers. Smooth torus Mesh adds 216 KiB of normals. Training-state and inference-model storage are reported separately. TransparentGS counts 53,048 foreground and 1,532,484 background Gaussians, not bytes; its mesh and probes are excluded.
\end{minipage}
\end{table}

\subsection{Synthetic scenes and baseline conditions}
\label{appendix:sec:supp-input-comparison}
\subsubsection{Layer count and fixed-inner capacity}
The comparison across layer counts covers $L=1,\ldots,6$ layers for both spherical shells and rotated eccentric ellipsoidal shells. For each scenario, a fixed set of 65 rays is used, corresponding to 13 local target offsets for each of the five camera positions. The camera positions are $(0,0,-3)$, $(1.25,0.2,-2.7)$, $(-1.15,-0.15,-2.7)$, $(0.2,1.2,-2.7)$, and $(-0.2,-1.15,-2.7)$. The reference target is the origin for the spherical shells and $(0.06,-0.04,0.02)$ for the ellipsoidal shells. Within the local horizontal/vertical plane for each viewpoint, the target offsets include $(0,0)$, $(\pm0.025,0)$, $(0,\pm0.025)$, $(\pm0.05,0)$, $(0,\pm0.05)$, and all four combinations of $(\pm0.035,\pm0.035)$. The background plane is set at $z=3$, and the primary transmission light path is budgeted for 64 interaction events. For each layer-count configuration, rays are re-traced starting from the camera, rather than simply truncating the initial segments of the light paths calculated for the six-shell configuration.

The rotated eccentric ellipsoidal shells are constructed by applying shell-specific translations, anisotropic scaling, and rotations to spherical meshes, with the geometric parameters listed below. Gaussian primitives are fitted to samples from the input meshes using local least squares, without directly using analytic curvature. The fixed per-interface capacity comparison uses 400 Gaussians and QEM meshes with a target of 2,044 faces per interface. Storage comparisons are based on actual geometry bytes.

\par\vspace{6pt}\noindent\begingroup\centering
\small\setlength{\tabcolsep}{3pt}
\begin{tabularx}{\columnwidth}{@{}lXrr@{}}
\toprule
Shell $s$ & Axis ratios & Yaw & Pitch \\
\midrule
0 & $(1,.85,.90)$ & $0^\circ$ & $0^\circ$ \\
1 & $(1.01,.865,.89)$ & $7^\circ$ & $-4^\circ$ \\
2 & $(.99,.835,.91)$ & $-9^\circ$ & $6^\circ$ \\
3 & $(1.015,.86,.885)$ & $11^\circ$ & $-7^\circ$ \\
4 & $(.985,.845,.92)$ & $-13^\circ$ & $5^\circ$ \\
5 & $(1.005,.87,.895)$ & $15^\circ$ & $-6^\circ$ \\
\bottomrule
\end{tabularx}
\par\endgroup\vspace{6pt}
\noindent\textit{Construction settings.} Reference radii are $1.00,0.92,\ldots,0.12$ from outside to inside; shell centers are $s(0.012,-0.008,0.004)$. Rotations about $y$ then $z$ and the listed scalings transform subdivision-6 icospheres into the source meshes. QEM uses an infinite maximum error and boundary weight 1.

The fixed total capacity configuration distributes 4,800 Gaussians across $2L$ interfaces, with each interface containing $2,400/L$ primitives; the mesh capacity is matched in terms of byte count. For the local capacity diagnostic shown in Figure~4(d), the two innermost interfaces are held constant while the outer shells are incrementally added; evaluations are performed using 128 fixed-orientation probes at each of the retained inner interfaces. The corresponding ``fixed capacity per interface'' comparison consistently employs 400 Gaussians per interface. This approach allows for the decoupling of effects stemming from changes in local representation capacity versus those arising from alterations to the inner surfaces themselves. This local diagnostic is based on a single timing run and is not intended to serve as a performance benchmark for repeated timing measurements.
\subsubsection{Capacity sweep and patch controls}
The six-shell rotated ellipsoid uses all 65 frozen rays for the reported method summaries. Five Gaussian capacities, $100,200,400,800,1600$ per interface, are paired with QEM targets of $508,1020,2044,4092,8188$ faces per interface. Both smooth- and face-normal Mesh variants are retained. The error curves use the 44 rays common to the plotted cohort across all 15 configurations; the aggregate means and failure counts use the complete 65-ray set. Section~\ref{appendix:sec:supp-capacity} provides the complete results.

The additional quadratic and cubic B\'ezier controls use the same source geometry, rays, media, and background. Source-projected radial samples define fixed triangulations with 512 vertices/1,020 faces and 280 vertices/556 faces per interface, respectively. Shared edge control points enforce $C^0$ continuity; $C^1$ continuity is not imposed.
\subsubsection{RefRef sphere: 100 test views}
The RefRef sphere uses the fixed 100-training/100-test split and native $512\times512$ test images. GaussTrans fits 400 cubic primitives to the training-only TransparentGS SDF mesh prepared by R3F, after conversion to the evaluation coordinate scale; it does not use the true source sphere. The same-input Mesh uses exactly this estimated geometry. Both share refractive index 1.5, the R3F background model frozen at step 9,999, 32 events, 64 nodes, and tail tolerance $10^{-3}$. No additional object-photo refinement, clean-background observation, or absorption prior is introduced.

The Gaussian query uses 128 coarse samples, at most 64 bisection iterations, root-distance and residual tolerances $10^{-6}$, support threshold 0.01, and gradient threshold 0.001. The normal-bandwidth factor is 0.5, normal support is three bandwidths, presence threshold is 0.001, and support-edge margin is 0.02. Background evaluation uses a seeded joint union of both methods' rays, with base seed 0 and chunks of 8,192; identical queries share identical background samples. All test views and hard failures are retained; Section~\ref{appendix:sec:supp-sphere} reports aggregate image metrics and transport bounds.

TransparentGS, TNSR, R3F, RT-Splatting, 3DGS, and 2DGS each retain their original training procedures and appearance modeling approaches. TransparentGS learns the refractive index, whereas TNSR fixes it at 1.5. TNSR employs two training stages, each consisting of 40,001 steps, while RT-Splatting is trained for 61,000 iterations. The reported results for RT-Splatting are initialized using SfM background points derived from the training images. Since the native inference models for these methods incorporate appearance components, they cannot be directly compared in terms of storage requirements against Gaussian-based representations that encompass only geometry. While the use of common test images facilitates comparisons of image quality, this does not imply any superiority in reconstruction performance under identical priors, nor does it suggest an advantage in runtime speed given the same computational budget.
\subsubsection{Torus and full Fresnel images}\label{appendix:sec:torus-fresnel}
The RefRef torus utilizes a non-convex geometry provided by the author, along with a background texture atlas generated from the lighting of the source scene and subsequently frozen. The viewpoint (view~30) is fixed before Gaussian scoring; rendering is performed at a native resolution of $512\times512$ with one sample per pixel. The total image consists of 262,144 pixels, of which 30,843 pixels correspond to the foreground region defined by the source mesh, and 6,069 pixels correspond to regions involving four interface crossings. These four crossings occur on a single non-convex object---following a medium sequence of air/glass/air/glass/air---rather than across four separate, nested shells. The model, composed of 400 Gaussians, remains frozen. The target face counts for QEM simplification are 1,376, 2,754, and 5,510, corresponding to approximately half, equal, and double the storage requirements, respectively, while the full source mesh is also retained. Each mesh variant includes both face normals and smoothed normals; the actual byte sizes for the core geometry and the additional normal buffers are listed separately. The fixed background atlases have a resolution of $256\times256$, independent of the output resolution; they do not reproduce all aspects of the source renderer's light visibility, antialiasing, and denoising.

We compare the nine configurations under identical background and tracing settings, using a budget of 32 events and 64 nodes with a tail tolerance of $10^{-3}$. To evaluate image accuracy for paths crossing four interfaces, we consider the corresponding region of 6,069 pixels. Since hard failures do not provide valid color predictions, paired RGB error bounds are computed on the 5,560 pixels that are free of hard failures across all nine configurations. The other 509 pixels are excluded only from this common-valid error comparison. Hard failures are counted separately for each method over the entire image, so a method cannot obtain a better error score simply by failing on more pixels.

The six-shell image in Figure~S1 instead uses the rotated eccentric ellipsoid scene with analytic, 400-Gaussian, 2,044-face smooth QEM, and full-source smooth Mesh arms. Its $512\times512$ camera is at $(1.25,0.2,-2.7)$, aimed at $(0.06,-0.04,0.02)$ with $45^\circ$ vertical field of view. It uses 128 events, 512 nodes, tail tolerance $10^{-4}$, and ray epsilon $10^{-6}$. Figure~S2 shows the torus images with all hard-failure pixels retained and marked.
Table~\ref{appendix:tab:supp-input-comparison} summarizes the method-specific inputs for the two sphere datasets. The detailed real-data protocol follows; different priors preclude an equal-prior reconstruction claim.
\setcounter{table}{8}
\begin{table*}[t]
\centering

\captionsetup{hypcap=false}
\caption{Comparison conditions. Native inference models can include appearance and background components; geometry-only bytes exclude those components and runtime caches (Section~\ref{appendix:sec:storage-accounting}). Adapted variants are identified explicitly.}
\label{appendix:tab:supp-input-comparison}
\centering\small\setlength{\tabcolsep}{4pt}\renewcommand{\arraystretch}{1.04}

\textit{(a) Synthetic RefRef sphere: main Table 2, 100 test views}\\[3pt]
\begin{tabular}{@{}>{\raggedright\arraybackslash}p{.18\linewidth}>{\raggedright\arraybackslash}p{.31\linewidth}>{\raggedright\arraybackslash}p{.28\linewidth}>{\raggedright\arraybackslash}p{.18\linewidth}@{}}
\toprule
Method & Geometry / RGB supervision & Medium / background & Accounting / scope\\
\midrule
GaussTrans; Mesh & Same estimated mesh; no extra RGB refinement & IOR 1.5; shared frozen background & Geometry only\\
TransparentGS & Native RGB-trained geometry and appearance & Learned IOR; native probes/background & Native models\\
TNSR & Native two-stage RGB refinement & IOR 1.5; native rendering & Inference model; no optimizer\\
R3F & Native mesh and RGB appearance & Native optical/background model & Native models\\
RT-Splatting & RGB training; training-SfM initialization & Native reflection/transmission model & Native models\\
3DGS; 2DGS & RGB-trained primitives & No shared refractive transport chain & Native Gaussian model\\
\bottomrule
\end{tabular}
\par\vspace{8pt}
\textit{(b) Captured Eikonal Fields Ball: main Figure 1, training view 19}\\[3pt]
\begin{tabular}{@{}>{\raggedright\arraybackslash}p{.18\linewidth}>{\raggedright\arraybackslash}p{.31\linewidth}>{\raggedright\arraybackslash}p{.28\linewidth}>{\raggedright\arraybackslash}p{.18\linewidth}@{}}
\toprule
Method & Geometry / RGB supervision & Medium / background & Accounting / scope\\
\midrule
GaussTrans & Contour sphere; five controls from 192 RGB samples & IOR 1.505; frozen background NeRF & Restricted calibration\\
3DGS; 2DGS & Shared sphere/SfM initialization; learnable geometry & Native learned appearance/background & Native losses\\
RT-Splatting & Shared sphere, mask and training-SfM initialization & Native reflection/transmission model & Final trained model\\
Eikonal Fields & Author-trained model; no local retraining & Native author field & Native evaluation split\\
TransparentGS & Shared sphere replaces geometry/normal proxy & Learned material and original probes & Adapted variant\\
R3F & Shared sphere as input mesh & Native optical and appearance model & Adapted variant\\
NU-NeRF & Shared sphere sets bounds; learned geometry & Native direct-interface model & Raw unmasked RGB; numerical fixes\\
\bottomrule
\end{tabular}

\end{table*}
\subsection{Captured Ball: Figure 1 protocol}
\label{appendix:sec:supp-real-ball}
Main-paper Figure~1 uses the captured Eikonal Fields Ball. Its training-view illustration is separate from the synthetic RefRef sphere test and is not a storage-matched or held-out reconstruction benchmark.

\subsubsection{Data, cameras, and displayed view}
This real-world dataset consists of 96 photographs along with their corresponding camera poses. Images at a resolution of $672\times504$ are used for local fitting and baseline training, while the reference images shown in Figure~1 are sampled from the original $4032\times3024$ photographs. Viewpoint~19---which is part of the training set---is used for the comparison; this specific viewpoint was selected from the set of views common to all evaluated methods before the zoomed-in panels were prepared, and it was not chosen based on error rankings.

The photograph and all method outputs in Figure~1 use the same object region, uniformly sampled into $512\times512$ images to keep the displayed area consistent. Camera settings and normalized scene coordinates follow the supplied configuration, without additional distortion calibration.

\subsubsection{GaussTrans input and local calibration}
The object is modeled as one solid, closed glass sphere. Its initial center and radius are estimated from automatic image contours, not from a CT scan or ground-truth mesh. A subdivision-4 icosphere supplies the input mesh for 400 cubic Gaussian primitives, with connected-neighborhood fitting and local validity domains. Intersections and normals are evaluated from the Gaussian-MLS field, using the fixed guided search and validation rules. The analytic sphere is not substituted for Gaussian intersections or normals.

The model in Figure~1 is initialized using a sphere estimated from the object's silhouette. Leveraging 192 RGB samples collected from four viewpoints, a five-parameter local calibration---covering the sphere's center, radius, and refractive index---is performed, after which the model is frozen. During this calibration process, the local surface coefficients of the Gaussian representation remain fixed, with only translation and uniform scaling applied to the entire interface.

\looseness=1
We use a bounded coordinate search with two step sizes and 21 objective evaluations, including initialization. Updates must reduce the training objective and satisfy the fixed acceptance criteria without increasing hard failures or budget truncations. The final candidate must improve the objective by at least 0.25\% over initialization.

\begin{table*}[t]
\centering\small\setlength{\tabcolsep}{5pt}
\caption*{Fixed settings for the local calibration and display in main-paper Figure~1.}
\begin{tabularx}{\textwidth}{@{}p{.23\textwidth}X@{}}
\toprule
Setting & Value \\
\midrule
Calibration observations & Views $19,57,76,95$, with 48 fixed RGB samples per view. \\
Objective & All-sample RGB MAE plus $0.1$ times the mean combined unresolved-color bound; failed samples remain included. \\
Controls and search order & Center offsets $(x,y,z)$, relative radius change, and IOR offset; search order: IOR, radius, $x$, $y$, $z$. Canonical coefficients and supports remain fixed. \\
Step sizes & Center offsets: $0.005$, then $0.0025$, normalized by initial radius; radius and IOR offsets: $0.01$, then $0.005$. \\
Parameter bounds & Center-offset norm $\leq1\%$ of initial radius; relative radius change $\leq2\%$; IOR in $[1.47,1.53]$. \\
Acceptance limits & No increase in hard-failure or budget-truncation counts. Relative to initialization, maximum increases are $10^{-4}$ in mean unresolved optical weight, $5\times10^{-4}$ in combined unresolved contribution, $0.002$ in any training-view MAE, and $0.5/1.0$ pixels in median/P90 contour residual. Each accepted update decreases the objective by more than $10^{-6}$. \\
Display region & ROI corners $(268.4002,88.5993)$ and $(527.2006,346.3457)$ in the $672\times504$ image; pixel centers are sampled uniformly. Scene coordinates are normalized, not metric. \\
Tracing budget & 12 events, 32 nodes, tail tolerance $0.005$; optical and background-opacity uncertainty are tracked separately. \\
\bottomrule
\end{tabularx}
\end{table*}

\looseness=1
The final selected normalized offsets are $(0.0025,-0.0075,0,0,0.005)$. The resulting sphere center is approximately $(-0.0136315,0.0073358,0.0738932)$, with a radius of $0.1623548$ and a refractive index of $1.505$. These values represent the scene model parameters obtained through fitting. Once the local fitting results are fixed, they are verified using viewpoint~0.

Background radiance is provided by a frozen NeRF queried outside the object's bounding box, while GaussTrans computes intersection and refraction at the glass interface. The image displays the resolved light-path contributions and retains all 262,144 pixels. Of these, 72 contain hard-failure branches; their colors are shown only as previews.

\subsubsection{Conditions of the displayed baselines}
\looseness=1
For the locally trained baselines, a set of evaluation viewpoints is fixed by selecting every eighth viewpoint---specifically, $\{0,8,\ldots,88\}$---while the remaining 84 viewpoints, including viewpoint~19, are used for training. The training images have a resolution of $672\times504$. The comparative methods follow the settings described below.

\paragraph{3DGS and 2DGS.}
Both methods use their native losses and 30,000 iterations, using the final trained model. Initialization contains 10,000 points on the shared estimated sphere and 18,697 background points retriangulated from training observations. Geometry remains learnable. Full RGB images and cameras are supplied; no refractive-index loss or training mask is added.

\paragraph{RT-Splatting.}
\looseness=1
The native method is trained for 61,000 iterations, using the estimated-sphere initialization, its projected mask, and training-retriangulated background points. The displayed output is from the fixed final iteration. Native $4032\times3024$ predictions are sampled at $\lfloor6\mathbf u\rfloor$, matching the photograph indices; their offset from continuous ROI rays is less than $1/6$ of a low-resolution pixel per coordinate.

\paragraph{Eikonal Fields.}
We evaluate the author-trained model with its native rendering mode, using 512 samples and Euler ODE integration. No local retraining is performed. Its settings specify every tenth view for evaluation, rather than every eighth view in the local split.

\paragraph{TransparentGS.}
The displayed variant replaces its learned geometry/normal proxy with the shared estimated sphere and uses its projected mask for optical support and background composition. Learned material parameters and original probes are retained without retraining for this replacement. Training uses 30,000 background, 7,000 coarse, 20,000 SDF, and 7,000 material iterations. The geometry-only variant was selected on training view~19 and then applied unchanged to the check views.

\paragraph{R3F.}
The shared-estimated-mesh variant uses 40,962 vertices and 81,920 faces, with 10,000 background and 10,000 foreground training iterations. Its geometry-production stage is replaced by the supplied estimated sphere. Native optical rendering is retained, and the final foreground model is evaluated.

\paragraph{NU-NeRF close-ups.}
\looseness=1
The right-hand close-ups show NU-NeRF's direct-interface variant. Training consists of 300,000 steps in stage~1, followed by 150,000 steps in stage~2 using the author's real-object direct-interface preset. The mesh obtained in stage~1 remains fixed during stage~2. The shared estimated sphere defines only the coordinate bounds and does not replace the learned geometry. We evaluate the final stage~2 model and display its RGB output before applying the native evaluation mask. Rays excluded by that mask remain visible, but their optical validity is not established. Numerical compatibility fixes are limited to using integer powers and explicitly specifying the cross-product axis; the optical equations remain unchanged.

\subsubsection{Display and interpretation}
The results from each method consistently display the same object region, with an output resolution of $512\times512$. These are derived from the respective native outputs via sampling or scaling; no method-specific adjustments---such as brightness modification, sharpening, or the replacement of photographic pixels---have been applied.

The two magnified regions on the right correspond to the following coordinate ranges within the ROI image: $[170,340)\times[140,310)$ and $[48,200)\times[176,328)$, where the origin is at the top-left corner and coordinates are expressed as $(x,y)$. The real-world photograph, GaussTrans, and NU-NeRF are all cropped to these same regions to illustrate textures and object boundaries affected by refraction; these crops are not intended for unbiased, localized quantitative evaluation. Figure~1 displays the appearance as captured in the real-world photograph under the given conditions; it does not serve as evidence of multi-layer reconstruction capability or demonstrate an advantage when using identical priors.

%% file: supplement/sections/b_results.tex
\setcounter{table}{1}
\begin{table*}[t]
\centering
\captionsetup{hypcap=false}
\caption{Six-shell capacity sweep. Counts are Gaussians or triangles per interface. Completed-path means use each configuration's completed paths; common-path means use the same 44 rays for all configurations. Timing is the median [minimum, maximum] of three repeats over all 65 rays. Geometry is uncompressed numerical storage; landing errors use scene units.}
\label{appendix:tab:supp-capacity}
\small
\setlength{\tabcolsep}{3pt}
\renewcommand{\arraystretch}{1.05}
\begin{tabular*}{\textwidth}{@{\extracolsep{\fill}}lrrrrlr@{}}
\toprule
Method & \shortstack{Count /\\interface} & \shortstack{Geometry\\(MiB)} & \shortstack{Landing\\completed} & \shortstack{Landing\\common 44} & Trace time (s) & \shortstack{Complete\\paths} \\
\midrule
GaussTrans & 100 & 0.2106 & 0.009102 & 0.009335 & 6.054 [6.043, 6.054] & 65/65 \\
GaussTrans & 200 & 0.4211 & 0.003033 & 0.002899 & 6.662 [6.640, 6.696] & 65/65 \\
GaussTrans & 400 & 0.8423 & 0.001498 & 0.001616 & 8.454 [8.442, 8.480] & 65/65 \\
GaussTrans & 800 & 1.6846 & 0.000878 & 0.000957 & 11.397 [11.340, 11.512] & 65/65 \\
GaussTrans & 1600 & 3.3691 & 0.000904 & 0.000996 & 18.769 [18.367, 19.505] & 65/65 \\
QEM (smooth) & 508 & 0.2098 & 0.186573 & 0.168674 & 0.357 [0.356, 0.357] & 65/65 \\
QEM (face) & 508 & 0.2098 & 3.176846 & 3.319414 & 0.251 [0.251, 0.251] & 48/65 \\
QEM (smooth) & 1020 & 0.4208 & 0.168965 & 0.163887 & 0.360 [0.359, 0.361] & 65/65 \\
QEM (face) & 1020 & 0.4208 & 1.497003 & 1.532664 & 0.292 [0.292, 0.296] & 62/65 \\
QEM (smooth) & 2044 & 0.8427 & 0.083899 & 0.081696 & 0.371 [0.370, 0.374] & 65/65 \\
QEM (face) & 2044 & 0.8427 & 0.776996 & 0.815376 & 0.305 [0.303, 0.306] & 62/65 \\
QEM (smooth) & 4092 & 1.6864 & 0.079477 & 0.083565 & 0.399 [0.394, 0.536] & 65/65 \\
QEM (face) & 4092 & 1.6864 & 0.557794 & 0.535102 & 0.333 [0.333, 0.334] & 65/65 \\
QEM (smooth) & 8188 & 3.3739 & 0.039761 & 0.037231 & 0.434 [0.433, 0.436] & 65/65 \\
QEM (face) & 8188 & 3.3739 & 0.371704 & 0.328926 & 0.370 [0.370, 0.371] & 65/65 \\
\bottomrule
\end{tabular*}
\end{table*}

\Needspace{10\baselineskip}
\section{Complete Quantitative and Image Results}
\label{appendix:sec:results}
\subsection{Capacity and tracing cost}
\label{appendix:sec:supp-capacity}
Table~\ref{appendix:tab:supp-capacity} expands the six-shell sweep in main-paper Section~4.2, reporting storage, landing error, tracing time, and completed-path counts. It distinguishes means over each method's completed paths from means over the 44 paths completed by every compared configuration.

Increasing Gaussian capacity from 100 to 800 reduces mean landing error from 0.009102 to 0.000878. At 1600 it rises to 0.000904 (3.0\%), while median tracing time increases by 64.7\%. Three interleaved CPU timing repeats use float64 and one Torch thread. GaussTrans takes 6.05--18.77 s versus 0.36--0.43 s for smooth-normal QEM, with matched-budget median-time ratios of 17.0--43.2$\times$. Face-normal QEM completes 48, 62, 62, 65, and 65 paths, respectively: 508 faces incur 16 TIR and one sequence deviation; 1020 and 2044 faces each incur three TIR.

\subsection{Sphere imaging and transport bounds}
\label{appendix:sec:supp-sphere}
In addition to the image-quality comparisons presented in Table~2 of the main text, Table~\ref{appendix:tab:supp-sphere-bounds} provides the upper and lower bounds of the mean absolute error (MAE) for object regions, the weights of unresolved light paths, and the number of hard failures. Both GaussTrans and Mesh were evaluated across all 100 test viewpoints at a resolution of $512\times512$. The metrics were calculated based on floating-point images---prior to their conversion into 8-bit format---and then averaged across all viewpoints. The upper and lower bounds here reflect the contribution of unresolved light paths rather than statistical confidence intervals; notably, unresolved contributions may persist even in the absence of hard failures.

\setcounter{table}{2}
\begin{table}[t]
\centering
\captionsetup{hypcap=false}
\caption{Sphere transport diagnostics, averaged over 100 views. Unresolved weights are full-image means. Hard failures are total pixel counts over all views.}
\label{appendix:tab:supp-sphere-bounds}
\small
\setlength{\tabcolsep}{2pt}
\begin{tabular*}{\columnwidth}{@{\extracolsep{\fill}}lrrr@{}}
\toprule
Method & Object MAE bounds & \shortstack{Mean\\unresolved} & \shortstack{Hard\\pixels} \\
\midrule
GaussTrans & [0.13897, 0.14265] & $7.67\!\times\!10^{-4}$ & 0 \\
Mesh & [0.13938, 0.14318] & $7.93\!\times\!10^{-4}$ & 0 \\
\bottomrule
\end{tabular*}
\end{table}

\subsection{Full Fresnel and non-convex\texorpdfstring{\\}{ }image comparisons}
\label{appendix:sec:supp-qualitative}
Figures~\ref{appendix:fig:six-shell-rendering} and~\ref{appendix:fig:method} supplement main-text Sections~4.2 and~4.4 with $512\times512$ renderings of the six-shell rotated eccentric ellipsoid and the RefRef torus at view~30, respectively. Within each scene, GaussTrans and Mesh share the input geometry, background, and tracing budget; the RGB-trained torus baselines retain their respective priors. The images show resolved color contributions under finite Fresnel budgets, including the analytic reference in Figure~\ref{appendix:fig:six-shell-rendering}. For the torus, paired RGB error bounds are computed over the common valid pixels in the four-crossing region, while hard failures are counted separately across the entire image and marked in the comparison. Detailed settings are provided in Section~\ref{appendix:sec:torus-fresnel}.
\setcounter{figure}{0}
\begin{figure*}[t]
  \centering
  \includegraphics[width=\textwidth]{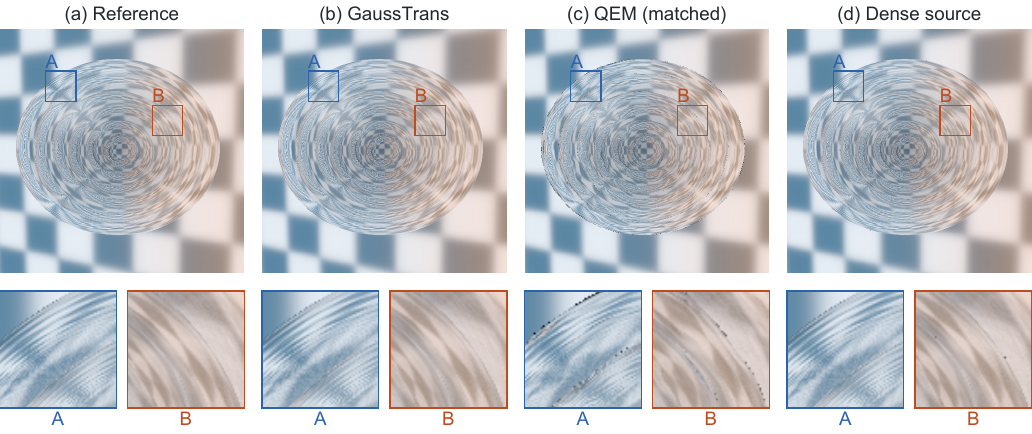}
  \caption{Six-shell transport. (a) Analytic reference; (b) GaussTrans; (c) storage-matched smooth QEM; (d) dense source mesh. Native $512\times512$ outputs and identical $64\times64$ detail crops A/B ($3.8\times$), without modifying the rendered pixels. All share a fixed Fresnel budget. Foreground accumulated-RGB MAE (b--d): 0.0100/0.0750/0.0067; geometry: 862.5/862.9/34,560.6 KiB. Mean unresolved foreground weight is approximately 5.3\% for (a,b,d): finite-budget images, not converged references.}
  \label{appendix:fig:six-shell-rendering}
\end{figure*}
\begin{figure*}[t]
  \centering
  \includegraphics[width=\textwidth]{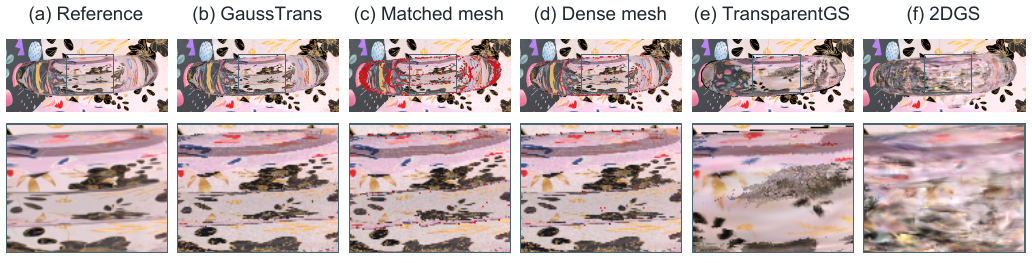}
  \caption{Torus, RefRef view 30. (a--f) Reference, GaussTrans, matched/dense smooth meshes, TransparentGS, 2DGS. Insets magnify the same box. (b--d) Supplied geometry, resolved RGB; core storage 96.88/96.82/648.00 KiB (normal caches excluded); hard failures 0/768/166 of 30,843 foreground pixels, marked red. RGB-trained (e,f) use different input priors.}
  \label{appendix:fig:method}
\end{figure*}

%% file: supplement/sections/c_ablations.tex
\Needspace{10\baselineskip}
\section{Extended Ablations}
\label{appendix:sec:ablations}
\subsection{Polynomial order and\texorpdfstring{\\}{ }within-primitive updates}
\label{appendix:sec:order-within}
\label{appendix:sec:local-fitting}
\label{appendix:sec:supp-refinement}
\paragraph{Polynomial order (main Table 1(a)).}
In the order-ablation study, all 12 interfaces---along with their 400 centers per interface, local coordinate systems, support scales, and 65 associated rays---are held fixed. The first-order case utilizes the tangent plane defined by the input mesh, with all higher-order coefficients set to zero; the second-order case involves re-fitting 3 coefficients; and the third-order case employs the 7 coefficients derived from the original joint fit. Higher-order fits follow the same construction method, utilizing the same input neighborhoods and normalized weighted least squares. For a surface of order $N$, the number of coefficients (excluding constant and linear terms) is $(N+1)(N+2)/2-3$; thus, orders two through eight involve 3, 7, 12, 18, 25, 33, and 42 coefficients, respectively. Fitting is performed using fixed Gaussian distance weights within the original tangential support radius. For a set of $M_N$ coefficients, the process requires at least $\max(16,M_N)$ samples, and additionally requires that the fitting matrix be full-rank and have a condition number no greater than $10^8$. In cases where a higher-order fit fails, no corrective measures---such as expanding the neighborhood or applying additional regularization---are employed.

For orders two through eight, runtime is measured on the same 65 rays after one warm-up, using three interleaved timing repeats. Guided intersection and scalar-evaluation optimizations are disabled. The first-order case is included only in the path-status comparison.
\setcounter{table}{3}
\begin{table*}[t]
\centering
\captionsetup{hypcap=false}
\caption{Six-shell optical-path results. Landing errors are in scene units, scaled by $10^3$; direction error is in degrees.}
\label{appendix:tab:supp-crossshell-paths}
\small
\setlength{\tabcolsep}{3pt}
\renewcommand{\arraystretch}{1.08}
\begin{tabular*}{\linewidth}{@{\extracolsep{\fill}}llrrrrr@{}}
\toprule
Input & Update & Complete & Mean & P90 & Maximum & Direction \\
\midrule
A & Initial & 39/39 & 11.884 & 20.719 & 42.697 & 0.1823 \\
A & Independent & 39/39 & 11.498 & 15.160 & 18.609 & 0.1857 \\
A & Coupled & 39/39 & 8.352 & 11.669 & 16.659 & 0.1336 \\
B & Initial & 39/39 & 12.015 & 21.499 & 34.862 & 0.1855 \\
B & Independent & 39/39 & 12.527 & 16.517 & 23.293 & 0.2029 \\
B & Coupled & 39/39 & 8.015 & 11.760 & 19.141 & 0.1295 \\
\bottomrule
\end{tabular*}

\end{table*}

\setcounter{table}{4}
\begin{table*}[t]
\centering
\captionsetup{hypcap=false}
\caption{Geometry and cost. Position: mean error in scene units scaled by $10^3$; normal: mean degrees. All 384 probes per configuration find roots. Times are seconds; search includes candidate work. Evaluation construction and tracing are measured separately.}
\label{appendix:tab:supp-crossshell-cost}
\small
\setlength{\tabcolsep}{3pt}
\renewcommand{\arraystretch}{1.08}
\begin{tabular*}{\linewidth}{@{\extracolsep{\fill}}llrrrrr@{}}
\toprule
Input & Update & Position & Normal & Search & Eval. build & Eval. trace \\
\midrule
A & Initial & 0.2230 & 0.0980 & 0.00 & 3.09 & 4.59 \\
A & Independent & 0.4260 & 0.1383 & 47.23 & 3.14 & 4.78 \\
A & Coupled & 0.3000 & 0.1046 & 48.25 & 1.53 & 4.54 \\
B & Initial & 0.2303 & 0.1067 & 0.00 & 3.23 & 4.95 \\
B & Independent & 0.4388 & 0.1537 & 47.94 & 3.20 & 5.02 \\
B & Coupled & 0.2870 & 0.1106 & 47.95 & 2.15 & 4.85 \\
\bottomrule
\end{tabular*}

\end{table*}

\paragraph{Within-primitive updates (main Table 1(b)).}
Two eccentric double-shell configurations, spherical and ellipsoidal, each contain four interfaces with 200 cubic Gaussians per interface. Both update variants receive the same coarse geometry, 16 synthetic background correspondences (eight through both shells and eight through the outer shell only), and two shell-shared deformation amplitudes. Search uses steps $0.02,0.01,0.005$, bounds $[-0.04,0.04]$, and 13 candidate evaluations per arm.

Centers follow the same fixed input-mesh sample indices. Center-only updating freezes frames and coefficients; coherent updating refits both from the deformed coarse mesh. The same controls produce identical center displacements; counts, supports, interface identities, cameras, and media remain fixed. Across the refinement comparisons, the primary-path objective is mean squared background-coordinate error, with cost $0.25$ per invalid path and regularizer $10^{-3}\|\boldsymbol\theta\|_2^2$. Failed paths stay in the denominator. Training observations alone select the final controls; full Fresnel imaging is evaluated separately. Evaluation comprises four fixed elevated-view rays, 24 geometric probes, and 512 dispersed pixels from two other views. Position and normal errors, interface sequences, TIR, and failures are retained separately. Analytic geometry supplies observations and final scores, not candidate updates.

\subsection{Cross-shell candidate selection}
\label{appendix:sec:supp-additional-ablations}
The six-shell ellipsoid ablation uses two frozen input perturbations (alternating and reversed), 400 cubic Gaussians per interface, six shell-shared controls, and ten synthetic correspondences from two cameras. Both arms perform coherent refitting and evaluate complete primary paths. Independent corrections hold all other shells at initialization during each shell's search, then assemble the selected corrections. The coupled arm uses the currently accepted other-shell controls in every candidate evaluation.

The independent and coupled strategies use the same objective, step sizes of $0.003$ and $0.0015$, and parameter bounds of $[-0.0045,0.0045]$. Each strategy receives 25 search evaluations, including initialization and rejected candidates, followed by one final training-set check. Updates are accepted only when the training objective improves, and the final result must outperform initialization under this objective. Evaluation uses 39 fixed rays from three cameras excluded from training and 384 geometric probes, with 32 probes per interface.

\paragraph{Two frozen inputs.}
All six configurations complete all 39 evaluation paths without hard failure, TIR, or event-budget termination. Coupled updates improve all 39 paired landing errors relative to independent updates in each input. Table~\ref{appendix:tab:supp-crossshell-paths} preserves the fixed denominator and reports the error tail. Table~\ref{appendix:tab:supp-crossshell-cost} separates search, evaluation-model construction, and evaluation tracing; these are single-run wall-clock measurements, not repeated timing estimates or comparisons with other experiments.

\paragraph{Geometric interpretation.}
Tables~\ref{appendix:tab:supp-crossshell-paths} and~\ref{appendix:tab:supp-crossshell-cost} show that coupled updates improve both geometric and optical-path accuracy relative to independent corrections. Compared with initialization, mean landing errors decrease by 29.7\% and 33.3\% for the two inputs, respectively, while mean position and normal errors increase. This reveals a trade-off between optical-path accuracy and local geometric accuracy under background-correspondence supervision.

\Needspace{5\baselineskip}
\paragraph{Per-interface geometry.}
Table~\ref{appendix:tab:supp-crossshell-interfaces} reports the mean errors for all 12 interfaces across both input sets. For each interface, 32 fixed geometric probes are defined, which are applicable across three configurations: initial, independently corrected, and coupled-update. These measurements serve to distinguish between improvements in the optical path and localized geometric changes.
\setcounter{table}{5}
\begin{table*}[t]
\centering
\captionsetup{hypcap=false}
\caption{Per-interface geometry for the fixed six-shell inputs. Position errors are scaled by $10^3$; normal errors are in degrees. I: initial, D: independent, C: coupled update. Interface identifiers denote the same physical surfaces across all three configurations.}
\label{appendix:tab:supp-crossshell-interfaces}
\small
\setlength{\tabcolsep}{3pt}
\renewcommand{\arraystretch}{0.95}
\begin{tabular*}{\linewidth}{@{\extracolsep{\fill}}llrrrrrr@{}}
\toprule
Input & Interface & Pos. I & Pos. D & Pos. C & Normal I & Normal D & Normal C \\
\midrule
A & 100 & 0.2686 & 0.7174 & 0.4181 & 0.0478 & 0.1216 & 0.0723 \\
A & 101 & 0.2686 & 0.7177 & 0.4182 & 0.0516 & 0.1319 & 0.0783 \\
A & 102 & 0.1896 & 0.2720 & 0.2720 & 0.0415 & 0.0580 & 0.0580 \\
A & 103 & 0.1888 & 0.2722 & 0.2722 & 0.0451 & 0.0634 & 0.0634 \\
A & 104 & 0.2324 & 0.6907 & 0.3852 & 0.0598 & 0.1632 & 0.0942 \\
A & 105 & 0.2321 & 0.6907 & 0.3849 & 0.0655 & 0.1828 & 0.1045 \\
A & 106 & 0.2683 & 0.1929 & 0.1929 & 0.0890 & 0.0662 & 0.0662 \\
A & 107 & 0.2675 & 0.1928 & 0.1928 & 0.1039 & 0.0770 & 0.0770 \\
A & 108 & 0.1700 & 0.6219 & 0.4711 & 0.0806 & 0.2828 & 0.2154 \\
A & 109 & 0.1703 & 0.6220 & 0.4713 & 0.1023 & 0.3622 & 0.2757 \\
A & 110 & 0.2104 & 0.0611 & 0.0611 & 0.1855 & 0.0588 & 0.0588 \\
A & 111 & 0.2097 & 0.0600 & 0.0600 & 0.3040 & 0.0914 & 0.0914 \\
B & 100 & 0.2284 & 0.6789 & 0.3784 & 0.0398 & 0.1144 & 0.0646 \\
B & 101 & 0.2276 & 0.6781 & 0.3779 & 0.0430 & 0.1241 & 0.0699 \\
B & 102 & 0.2824 & 0.1797 & 0.1797 & 0.0588 & 0.0383 & 0.0383 \\
B & 103 & 0.2810 & 0.1801 & 0.1801 & 0.0638 & 0.0427 & 0.0427 \\
B & 104 & 0.1990 & 0.6586 & 0.3522 & 0.0506 & 0.1554 & 0.0854 \\
B & 105 & 0.1984 & 0.6582 & 0.3515 & 0.0564 & 0.1754 & 0.0959 \\
B & 106 & 0.1614 & 0.2993 & 0.1466 & 0.0587 & 0.0972 & 0.0496 \\
B & 107 & 0.1614 & 0.2986 & 0.1461 & 0.0679 & 0.1140 & 0.0577 \\
B & 108 & 0.2746 & 0.7280 & 0.5768 & 0.1286 & 0.3357 & 0.2663 \\
B & 109 & 0.2740 & 0.7275 & 0.5762 & 0.1639 & 0.4315 & 0.3417 \\
B & 110 & 0.2375 & 0.0894 & 0.0894 & 0.2071 & 0.0822 & 0.0822 \\
B & 111 & 0.2382 & 0.0894 & 0.0894 & 0.3414 & 0.1335 & 0.1335 \\
\bottomrule
\end{tabular*}

\end{table*}

\setcounter{table}{6}
\begin{table*}[t]
\centering

\captionsetup{hypcap=false}
\caption{Sixteen-input refinement. Rows average case-level metrics; P90 is the mean of case-level P90s. Landing and position errors are scaled by $10^3$; normals are degrees. Complete paths use summed fixed denominators. Search time includes refitting under concurrent execution. The larger evaluation set is specified in Section~\ref{appendix:sec:supp-refinement16}.}
\label{appendix:tab:supp-refinement16}
\centering\small\setlength{\tabcolsep}{4pt}\renewcommand{\arraystretch}{0.95}
\begin{tabular*}{\linewidth}{@{\extracolsep{\fill}}llrrrrrr@{}}
\toprule
Input group & Update & Complete & Landing & P90 & Position & Normal & Search (s)\\
\midrule
Ellipsoid, base ($n=2$) & Initial & 294/294 & 14.059 & 21.674 & 0.227 & 0.102 & --\\
 & Center-only & 294/294 & 18.476 & 35.808 & 0.282 & 0.113 & 137.5\\
 & Independent & 294/294 & 12.545 & 19.446 & 0.433 & 0.145 & 146.8\\
 & Coupled & 294/294 & 8.315 & 12.805 & 0.294 & 0.107 & 146.0\\
\addlinespace[2pt]
Ellipsoid, matched bump ($n=3$) & Initial & 441/441 & 5.892 & 9.585 & 0.107 & 0.049 & --\\
 & Center-only & 441/441 & 6.527 & 11.285 & 0.131 & 0.055 & 140.6\\
 & Independent & 441/441 & 5.892 & 9.585 & 0.107 & 0.049 & 148.9\\
 & Coupled & 441/441 & 4.816 & 8.872 & 0.250 & 0.079 & 147.1\\
\addlinespace[2pt]
Ellipsoid, shifted bumps ($n=3$) & Initial & 441/441 & 6.632 & 11.664 & 0.029 & 0.028 & --\\
 & Center-only & 441/441 & 7.835 & 14.701 & 0.088 & 0.039 & 139.1\\
 & Independent & 441/441 & 6.632 & 11.664 & 0.029 & 0.028 & 146.2\\
 & Coupled & 441/441 & 6.715 & 12.015 & 0.137 & 0.048 & 147.6\\
\addlinespace[2pt]
Quartic, base ($n=2$) & Initial & 294/294 & 15.572 & 24.337 & 0.239 & 0.125 & --\\
 & Center-only & 294/294 & 29.129 & 54.330 & 0.321 & 0.139 & 125.5\\
 & Independent & 294/294 & 26.148 & 57.906 & 0.442 & 0.159 & 135.3\\
 & Coupled & 294/294 & 20.034 & 43.174 & 0.324 & 0.145 & 136.4\\
\addlinespace[2pt]
Quartic, matched bump ($n=3$) & Initial & 441/441 & 10.014 & 16.981 & 0.123 & 0.087 & --\\
 & Center-only & 441/441 & 14.070 & 29.006 & 0.239 & 0.104 & 125.4\\
 & Independent & 441/441 & 10.014 & 16.981 & 0.123 & 0.087 & 135.2\\
 & Coupled & 441/441 & 10.974 & 23.325 & 0.268 & 0.114 & 134.4\\
\addlinespace[2pt]
Quartic, shifted bumps ($n=3$) & Initial & 441/441 & 8.922 & 15.768 & 0.047 & 0.072 & --\\
 & Center-only & 441/441 & 14.192 & 33.037 & 0.145 & 0.084 & 127.7\\
 & Independent & 441/441 & 9.198 & 17.027 & 0.088 & 0.078 & 135.5\\
 & Coupled & 441/441 & 11.935 & 25.598 & 0.236 & 0.099 & 135.8\\
\bottomrule\end{tabular*}
\end{table*}

\setcounter{table}{7}
\begin{table*}[t]
\centering

\captionsetup{hypcap=false,skip=3pt}
\caption{Guidance on/off with matched bracket precision ($10^{-14}$). (a) Each view contains 256 pixels; hard failures, unresolved weight $U$, reference status/sequence agreement, and RGB bounds ($\times10^3$) on $n_{\rm RGB}$ common-valid pixels. UR/LL cameras: $(0.43,0.19,-3)$ and $(-0.21,-0.41,-3)$, aimed at the origin with $45^\circ$ vertical field of view. (b) Median [minimum, maximum] of three warmed serial passes over four UR pixels per shape, with query counters enabled; these are diagnostic timings, not full-image times.}
\label{appendix:tab:supp-guided1024}
\centering\small\setlength{\tabcolsep}{4pt}\renewcommand{\arraystretch}{0.95}
\begin{tabular*}{\linewidth}{@{\extracolsep{\fill}}llrrrrr@{}}
\toprule
(a) Shape/view & Guidance & Hard & $U$ (\%) & Status/sequence & $n_{\rm RGB}$ & RGB error bounds\\
\midrule
Ellipsoid/UR & Off & 0/256 & 2.42202 & 256/256 & 256 & [0.57532, 26.85211]\\
Ellipsoid/UR & On & 0/256 & 2.42202 & 256/256 & 256 & [0.57532, 26.85214]\\
Ellipsoid/LL & Off & 0/256 & 2.36166 & 255/256 & 256 & [1.24324, 26.98572]\\
Ellipsoid/LL & On & 0/256 & 2.36167 & 256/256 & 256 & [1.24293, 26.98559]\\
Quartic/UR & Off & 2/256 & 4.10307 & 254/256 & 254 & [17.52729, 70.88902]\\
Quartic/UR & On & 2/256 & 4.07761 & 254/256 & 254 & [17.54268, 70.74633]\\
Quartic/LL & Off & 1/256 & 4.22104 & 254/256 & 255 & [13.71367, 63.82558]\\
Quartic/LL & On & 1/256 & 4.26139 & 254/256 & 255 & [13.71330, 64.23462]\\
\bottomrule\end{tabular*}
\par\vspace{5pt}
\begin{tabular*}{\linewidth}{@{\extracolsep{\fill}}lrr@{}}
\toprule
(b) Shape & Guidance off (s) & Guidance on (s)\\
\midrule
Ellipsoid & 2.908 [2.890, 3.051] & 15.772 [15.771, 15.882]\\
Quartic & 3.662 [3.637, 3.694] & 13.183 [13.103, 13.377]\\
\bottomrule\end{tabular*}
\end{table*}

\subsection{Sixteen-input refinement extension}
\label{appendix:sec:supp-refinement16}
To test whether refinement gains depend on the initial geometric errors, we evaluate 16 perturbed inputs across six-shell ellipsoidal and quartic geometries (Table~\ref{appendix:tab:supp-refinement16}). For each input, we compare initialization, center-only updating, independent correction, and coupled updating. The three update strategies start from the same initial geometry and use identical training observations, deformation controls, objectives, and a budget of 26 objective evaluations. After refinement, we evaluate background landing errors using 147 rays from three viewpoints excluded from training, and measure surface position and normal errors using 576 geometric probes. This comparison examines whether the updates improve light propagation across different initial conditions and whether those improvements are accompanied by more accurate geometry.

All four configurations complete the prescribed optical paths and geometric evaluations across all 16 inputs. Compared with the other update strategies, coupled updating achieves lower mean landing errors in 11 cases against independent correction and 13 cases against center-only updating. Its improvements over the initial model are less consistent: landing error decreases in 9 cases, while no case simultaneously improves mean surface-position and normal errors. These results show that coupled updating often provides better optical-path accuracy than the alternative update strategies, but does not consistently improve optical and geometric accuracy together.

\subsection{Targeted guided-intersection diagnostics}
\label{appendix:sec:targeted-query}
We compare the intersection results obtained with and without guidance enabled, while keeping the Gaussian field and the intersection acceptance criteria constant. The experiments utilize two sets of fixed samples. The first set consists of 8 pixels---from a six-shell scene---where previous attempts failed, along with 4 control pixels that yielded successful results. The second set involves applying 41 distinct directional perturbations to a single grazing ray; by densely sampling the same field, we identified roots in 33 of these queries, which then serve as the basis for evaluating intersection recovery. Guidance is activated only when a ray is likely to enter the tangential support range of a primitive, and either the coarse search fails to locate a medium-compatible intersection, or fewer than two sample points are found within the support interval preceding a detected intersection. The image-tracing budget is set to 128 events and 512 nodes, with a tail tolerance of $10^{-4}$.

With the same image-tracing budget, enabling guidance reduced the number of hard failures among the 8 pre-selected problematic pixels from 8 to 2, while the average unresolved weight increased from 0.06317 to 0.06464. None of the 4 control pixels (which functioned normally) exhibited any hard failures. This indicates that guidance can help mitigate hard failures in certain challenging queries. Among the 33 grazing queries with observed roots, guidance increases intersection recovery from 20 to 33.

\subsection{Full-image guided-intersection ablation}
\label{appendix:sec:supp-guided1024}
To assess guided intersection in full-image rendering, we compare guidance enabled and disabled on six-shell ellipsoidal and quartic scenes. Two fixed viewpoints per scene are rendered at $16\times16$, giving 1,024 pixels in total. The two modes share the same geometry, cameras, Gaussian field, root-refinement precision, and intersection acceptance criteria; only candidate guidance differs. Both use a tracing budget of 128 events and 512 nodes, with a tail tolerance of $10^{-4}$. Table~\ref{appendix:tab:supp-guided1024} reports image errors, unresolved contributions, hard failures, and computational cost, retaining all pixels in the failure statistics.

Guidance recovers valid intersections on the selected difficult queries in Section~\ref{appendix:sec:targeted-query}, enabling surface-normal evaluation and subsequent reflection or refraction when coarse sampling misses an interface. In the 1,024-pixel evaluation, hard-failure counts remain unchanged, while one additional pixel matches the reference path status and interface sequence. The RGB error intervals overlap. These results support guidance for maintaining propagation through difficult intersections.

On four fixed pixels per scene, guidance increases the median tracing time from 2.908 to 15.772~s for the ellipsoidal scene and from 3.662 to 13.183~s for the quartic scene (Table~\ref{appendix:tab:supp-guided1024}). Construction and loading are excluded; these are not full-image rendering times.

All 1,024 pixels enter the failure counts. RGB errors use the common valid pixels of both configurations and the reference, excluding invalid hard-failure colors. The intervals bound unresolved light-transport contributions, not statistical confidence.